\documentclass{article}

\PassOptionsToPackage{numbers, compress}{natbib}
\usepackage[preprint]{neurips_2025}
\usepackage{fix-cm}

\usepackage{xcolor}         
\definecolor{linkColor}{rgb}{0.2,0.4,0.6}
\usepackage[utf8]{inputenc} 
\usepackage[T1]{fontenc}    
\usepackage[colorlinks=true,linkcolor=linkColor,citecolor=linkColor,filecolor=linkColor,urlcolor=linkColor]{hyperref}       
\usepackage{url}  
\usepackage{booktabs}       
\usepackage{amsfonts}       
\usepackage{nicefrac}       
\usepackage{microtype}      
\usepackage{wrapfig}

\usepackage{graphicx}
\usepackage{arydshln}
\usepackage{booktabs}
\usepackage{multirow}
\usepackage{caption}
\usepackage{subcaption}
\usepackage{makecell}
\usepackage{csquotes}
\usepackage{epigraph}
\usepackage{tcolorbox}
\tcbuselibrary{listings, breakable, skins}
\usepackage{caption}
\usepackage{alltt}

\RequirePackage{algorithm}
\RequirePackage{algorithmic}

\usepackage{multirow}
\usepackage{amsmath}
\usepackage{capt-of}
\usepackage{tabularx}
\usepackage{epsfig}
\usepackage{amssymb}
\usepackage{amsfonts}
\usepackage{booktabs}
\usepackage{scalerel}
\usepackage[inline]{enumitem}
\usepackage{listings}
\usepackage{varwidth}
\usepackage[export]{adjustbox}
\usepackage{tikz}
\usetikzlibrary{tikzmark}

\usepackage{stmaryrd}
\usepackage{bbm}
\usepackage{wrapfig}
\usepackage{pifont}
\usepackage[noabbrev]{cleveref}

\definecolor{deepblue}{rgb}{0,0,0.5}
\definecolor{officeblue}{RGB}{0,102,204}
\definecolor{deepred}{rgb}{0.6,0,0}
\definecolor{deepgreen}{rgb}{0,0.5,0}
\definecolor{mybrickred}{RGB}{182,50,28}

\definecolor{fillcolor}{RGB}{216,217,252}

\usepackage{etoolbox}
\usepackage{framed}

\newif\ifxetexorluatex
\ifxetex
  \xetexorluatextrue
\else
  \ifluatex
    \xetexorluatextrue
  \else
    \xetexorluatexfalse
  \fi
\fi
\newcommand*\quotesize{60} 
\newcommand*{\openquote}
   {\tikz[remember picture,overlay,xshift=-4ex,yshift=-2.5ex]
   \node (OQ) {\fontsize{\quotesize}{\quotesize}\selectfont``};\kern0pt}

\newcommand*{\closequote}[1]
  {\tikz[remember picture,overlay,xshift=4ex,yshift={#1}]
   \node (CQ) {\fontsize{\quotesize}{\quotesize}\selectfont''};}

\colorlet{shadecolor}{white}

\newcommand*\shadedauthorformat{\emph} 

\newcommand*\authoralign[1]{%
  \if#1l
    \def\authorfill{}\def\quotefill{\hfill}
  \else
    \if#1r
      \def\authorfill{\hfill}\def\quotefill{}
    \else
      \if#1c
        \gdef\authorfill{\hfill}\def\quotefill{\hfill}
      \else\typeout{Invalid option}
      \fi
    \fi
  \fi}
{\authoralign{#1}
\ifblank{#2}
   {\def\shadequoteauthor{}\def\yshift{-2ex}\def\quotefill{\hfill}}
   {\def\shadequoteauthor{\par\authorfill\shadedauthorformat{#2}}\def\yshift{2ex}}
\begin{snugshade}\begin{quote}\openquote}
{\shadequoteauthor\quotefill\closequote{\yshift}\end{quote}\end{snugshade}}

\newfloat{Code}{htbp}{Code}

\usepackage{amsmath,amsfonts,bm}

\def\eqref#1{equation~\ref{#1}}

\def\1{\bm{1}}

\DeclareMathAlphabet{\mathsfit}{\encodingdefault}{\sfdefault}{m}{sl}
\SetMathAlphabet{\mathsfit}{bold}{\encodingdefault}{\sfdefault}{bx}{n}

\newcommand\our{\makebox{\textsc{Rrm}}}
\newcommand\bsl{DirectJudge}

\title{Reward Reasoning Model}

\author{
Jiaxin Guo\thanks{~Equal contribution. $\diamond$ Corresponding author.}$~~^{1,2}$~~~~~Zewen Chi\footnotemark[1]$~~^{1}$~~~~~Li Dong\footnotemark[1]$~~^{1}$ \\
~\bf Qingxiu Dong$^{1,3}$~~~~~\bf Xun Wu$^1$~~~~~Shaohan Huang$^1$~~~~~Furu Wei$^1$$^{\diamond}$ \\
~$^1$ Microsoft Research ~~~~~
~$^2$ Tsinghua University ~~~~~
~$^3$ Peking University\\
~{\href{https://aka.ms/GeneralAI}{https://aka.ms/GeneralAI}}
}

\begin{document}

\maketitle

\begin{abstract}
Reward models play a critical role in guiding large language models toward outputs that align with human expectations. However, an open challenge remains in effectively utilizing test-time compute to enhance reward model performance. In this work, we introduce Reward Reasoning Models (\our{}s), which are specifically designed to execute a deliberate reasoning process before generating final rewards. Through chain-of-thought reasoning, \our{}s leverage additional test-time compute for complex queries where appropriate rewards are not immediately apparent. To develop \our{}s, we implement a reinforcement learning framework that fosters self-evolved reward reasoning capabilities without requiring explicit reasoning traces as training data. Experimental results demonstrate that \our{}s achieve superior performance on reward modeling benchmarks across diverse domains. Notably, we show that \our{}s can adaptively exploit test-time compute to further improve reward accuracy. The pretrained reward reasoning models are available at \url{https://huggingface.co/Reward-Reasoning}.
\end{abstract}

\vfill{}

\begin{figure}[h]
\centering
\includegraphics[width=\linewidth]{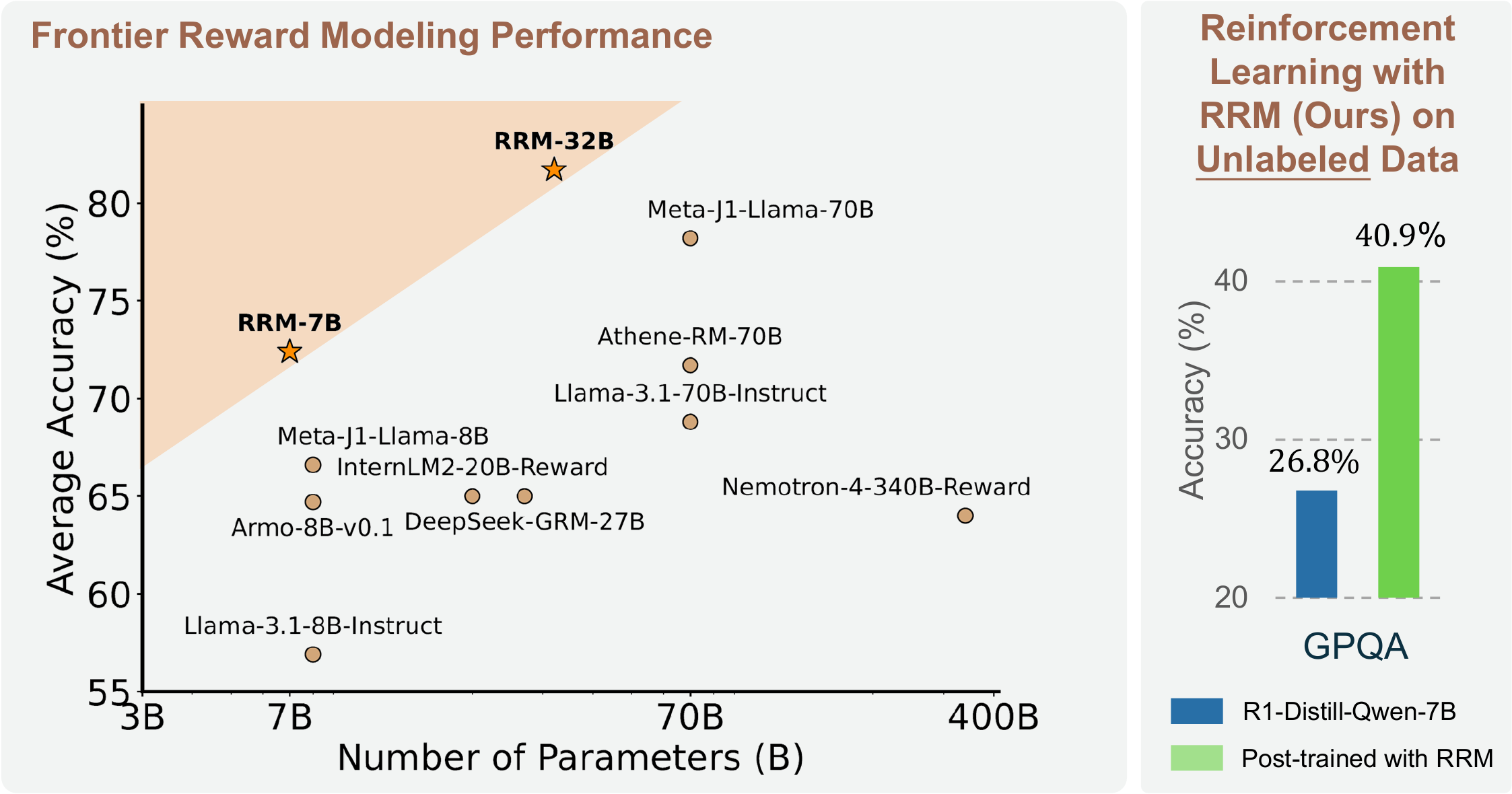}
\caption{Average accuracy of various reward models on Preference Proxy Evaluations~\cite{frick2025how} over the MMLU-Pro, MATH, and GPQA subsets. The proposed reward reasoning model (RRM) outperforms previous reward models across model sizes. We also conduct reinforcement learning on unlabeled data, using RRM as the reward model. Even without ground-truth answers, reinforcement learning with RRM achieves significant improvements on GPQA, which evaluates general-domain reasoning.}
\label{fig:scatter}
\end{figure}

\newpage

\section{Introduction}

Large language models (LLMs) such as GPTs \citep{gpt3,achiam2023gpt} have significantly transformed the field of artificial intelligence. In recent years, the development paradigm of LLMs has evolved from primarily scaling pre-training resources to emphasizing post-training techniques, driven by the dual imperatives of aligning models with human preferences \citep{gpt35} and enhancing specific capabilities like reasoning \citep{bai2022training, touvron2023llama}. This shift reflects a growing recognition that model performance depends not only on scale but also on sophisticated methods to refine model behavior after initial training.

Reinforcement learning has emerged as a fundamental approach in LLM post-training, leveraging supervision signals from either human feedback (RLHF) or verifiable rewards (RLVR) \citep{gpt35, 10.5555/3294996.3295184, gao2024designing, lambert2024t, guo2025deepseek}. While RLVR has shown promising results in mathematical reasoning tasks, it is inherently constrained by its reliance on training queries with verifiable answers~\citep{guo2025deepseek}. This requirement substantially limits RLVR's application to large-scale training on general-domain queries where verification is often intractable \citep{cobbe2021training, jimenez2024swebench, wang2024secrets}. In contrast, RLHF typically employs a reward model as a proxy for human preference, enabling more extensive application across diverse domains \citep{bai2022constitutional, ouyang2022training}. Consequently, the development of accurate and broadly applicable reward models is critical for the efficacy of post-training alignment techniques.

Recent work on reward models can be categorized into scalar reward models \citep{gpt35,liu2024skywork} and generative reward models \citep{chen2025judgelrm, skyworkcritic2024, DBLP:conf/iclr/WangYYZYW0J000024, zhu2025judgelm}. Scalar reward models typically replace the decoding layer with a linear head to predict a single scalar value. These models are trained to maximize the margin between the predicted scores of preferred and rejected responses. Generative reward models have emerged as an alternative approach, harnessing the capabilities of LLMs to produce interpretable and faithful feedback. These models offer enhanced flexibility, enabling them to follow adaptive evaluation instructions to construct synthetic training data, thereby facilitating self-improvement through iterative refinement \citep{gu2024survey, llm-as-a-judge}.

Despite the widespread application of current reward models, it remains an open challenge to effectively scale test-time compute for reward estimation. To serve as general-purpose evaluators, reward models should be capable of adapting to a diverse spectrum of queries, ranging from immediately obvious questions to complex tasks that require extensive reasoning \citep{10.5555/3618408.3618845, rafailov2024scaling}. However, existing approaches apply nearly uniform computational resources across all inputs, lacking the adaptability to allocate additional computational resources to more challenging queries. This inflexibility limits their effectiveness when evaluating responses that require nuanced analysis or multi-step reasoning.

To address the aforementioned challenge, we propose Reward Reasoning Models (\our{}s). Unlike existing reward models, \our{} frames reward modeling as a reasoning task, wherein the model first produces a long chain-of-thought reasoning process before generating the final rewards. Since supervised data providing reward reasoning traces are not readily available, we develop a training framework called Reward Reasoning via Reinforcement Learning, which encourages \our{}s to self-evolve their reward reasoning capabilities within a rule-based reward environment.
Furthermore, we introduce multi-response rewarding strategies, including the ELO rating system \citep{elo1978rating} and knockout tournament, enabling \our{}s to flexibly allocate test-time compute to practical application scenarios.

Extensive experiments on reward modeling benchmarks show that \our{}s consistently outperform strong baselines across multiple domains, including reasoning, general knowledge, safety, and alignment with human preference. Besides, we demonstrate the effectiveness of \our{}s by applying them in practical applications, specifically reward-guided best-of-N inference and post-training LLMs with \our{} feedback. More significantly, we conduct systematic analysis of the test-time scaling behaviors of \our{}s, revealing their capacity to adaptively utilize test-time compute to achieve enhanced performance. Furthermore, our analysis reveals that \our{}s develop distinct reasoning patterns compared to untrained foundation models, suggesting that our Reward Reasoning via Reinforcement Learning framework successfully guides models to develop effective reward evaluation capabilities. These insights provide deeper understanding of reward reasoning processes and will likely inspire the development of future reward reasoning models within the research community.

Our main contributions are as follows: 
\begin{itemize}[leftmargin=*]
\setlength\itemsep{0.01em}
\item We propose Reward Reasoning Models (\our{}s), which perform explicit reasoning before producing final rewards. This reasoning phase enables \our{}s to adaptively allocate additional computational resources when evaluating responses to complex tasks. \our{}s introduce a novel dimension for enhancing reward modeling by effectively scaling test-time compute, while maintaining general applicability and effectiveness across diverse evaluation scenarios.
\item We develop a framework named Reward Reasoning via Reinforcement Learning. This framework encourages \our{}s to self-evolve reward reasoning capabilities without requiring explicit reasoning traces as training data.
\item We conduct extensive experiments demonstrating not only the remarkable performance of \our{}s in reward modeling but also their promising test-time scaling properties.
\end{itemize}

\section{Related Work}
\label{sec:pre}

\paragraph{Reward Models}
Reward models can be characterized along two dimensions: reward formulation and scoring scheme \citep{ouyang2022training, zhong2025comprehensive}. Formulation strategies include numeric only, which assigns scalar scores to query-response pairs \citep{gpt35, liu2024skywork, wang2023helpsteer, wang2024helpsteer}, and generative, which produces natural language feedback from which rewards may be extracted \citep{alexandru2025atla, arabzadeh-etal-2024-assessing, cao2024compassjudger, chen2025judgelrm, liu2025inference, vu-etal-2024-foundational, ye2025learning, zhang2024generative}. Scoring schemes typically follow either absolute approaches, evaluating individual query-response pairs independently \citep{cobbe2021training, 10.5555/3618408.3618845, 10.1145/3701551.3703583, winata2025metametrics, yu-etal-2025-self, 10.5555/3692070.3694459}, or discriminative methods that compare candidate responses to express relative preferences \citep{jiang-etal-2023-llm, li2024generative, liu2025pairwise, park-etal-2024-offsetbias, skyworkcritic2024, wang2024self, zhu2025judgelm}.

\paragraph{Generative Reward Models}
Generative reward models (GRMs), conceptually aligned with the LLM-as-a-Judge paradigm \citep{wu2024meta, zheng2023judging}, offer nuanced, interpretable feedback with flexibility for both single-instance evaluation and multi-response comparison \citep{kim-etal-2024-prometheus, mahan2024generative}. This approach addresses limitations of traditional evaluation methods like ROUGE \citep{lin2004rouge} and BLEU \citep{papineni2002bleu}, which struggle with open-ended tasks requiring sophisticated judgment \citep{schluter2017limits}. GRMs can support judgment across diverse tasks, including multimodal inputs \citep{10203359, li2024generative, zhu2025judgelm}, and contemporaneous work on GRMs demonstrates promising scalability in both model capacity and inference-time compute \citep{chen2025rm, liu2025inference}. However, concerns persist about evaluation reliability, as LLMs may produce biased or hallucinated judgments that diverge from human standards \citep{achiam2023gpt, bubeck2023sparks}.

\paragraph{Inference-Time Scaling}

Inference-time scaling dynamically adjusts computational resources during model inference based on input complexity, inspired by human adaptive reasoning \citep{kaplan2020scaling, snell2025scaling, wu2025inference}. Recent approaches include parallel scaling strategies such as multi-sampling \citep{brown2024large} and reward model-guided aggregation \citep{lightman2024lets, snell2025scaling, prmlessons}, which combine multiple outputs to enhance quality. Alternative methods utilize horizon-based scaling to extend reasoning traces \citep{wei2022chain}. Advanced systems like OpenAI's o1 and DeepSeek's R1 series demonstrate spontaneous computational allocation that adjusts ``thinking horizons'' in response to task complexity \citep{guo2025deepseek, jaech2024openai}. These approaches collectively underscore the importance of inference-time adaptability in improving model performance, particularly on complex reasoning tasks.

\section{Reward Reasoning Model}
\label{sec:method}

\subsection{Input Representation}
\begin{figure}[t]
    \centering
    \includegraphics[width=1.0\linewidth]{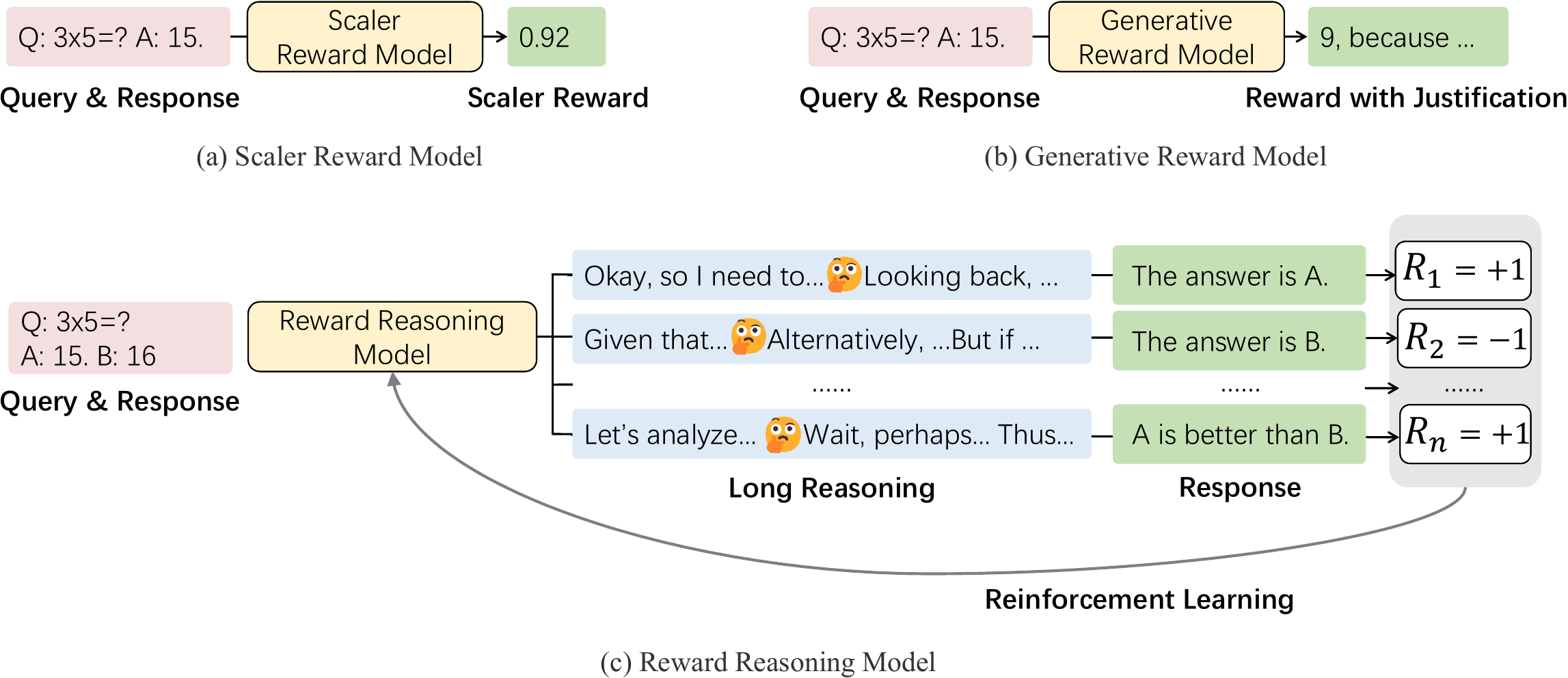}
    \caption{An overview of reward reasoning model (\our{}). \our{} adaptively leverages test-time compute through chain-of-though reasoning before producing rewards.}
    \label{fig:method}
\end{figure}

Figure~\ref{fig:method} provides an overview of reward reasoning models (\our{}s).
\our{}s utilize the Qwen2 \citep{qwen2} model architecture with a Transformer-decoder as backbone. We formulate the reward modeling task as a text completion problem, wherein \our{}s take queries and corresponding responses as input, and autoregressively generate output text consisting of a thinking process followed by a final judgment. Unlike existing reward models, \our{}s perform chain-of-thought reasoning before producing rewards, enabling them to leverage test-time compute adaptively. We refer to this process as reward reasoning.

Each input of \our{}s contains a query and two corresponding responses. The goal of \our{}s is to determine which response is preferred, with ties not allowed. We employ the system prompt from the RewardBench repository\footnote{\url{https://github.com/allenai/reward-bench}}, which guides the model to perform a systematic analysis of the two responses according to several evaluation criteria, including instruction fidelity, helpfulness, accuracy, harmlessness, and level of detail. The model is also explicitly instructed to avoid common biases (such as response order or length) and must justify its judgment through structured reasoning before outputting its final decision in the format `\textbackslash boxed\{Assistant 1\}' or `\textbackslash boxed\{Assistant 2\}'. The detailed prompt template is provided in Appendix~\ref{app:prompt:template}.

The input of \our{}s is restricted to exactly two candidate responses, thereby reserving output length capacity for reward reasoning. Section~\ref{subsec:multi_response} introduces methods by which \our{}s assign rewards to scenarios involving multiple candidate responses for a given query.

\subsection{Model Training with Reinforcement Learning}
We develop a training framework called Reward Reasoning via Reinforcement Learning to train \our{}s. Unlike conventional supervised fine-tuning approaches, which relies on existing reasoning traces, our framework encourages \our{}s to self-evolve their reasoning capacities within a rule-based reward environment. The reward function is defined as follows:
\begin{equation}
    \begin{aligned}
        \mathcal{R} =
    \begin{cases}
    +1, & \text{\our{} selects correct response} \\
    -1, & \text{otherwise}
    \end{cases}
    \end{aligned}
\end{equation}
Note that the reward $\mathcal{R}$ evaluates whether \our{} correctly prefers the ground-truth response, rather than scoring its own outputs. Despite the simplicity of the reward signals, such rule-based rewards can effectively supervise the policy models to develop reasoning patterns that lead to correct final judgments.

We use Deepseek-R1 distilled models \citep{guo2025deepseek} as base models, applying group relative policy optimization (GRPO) \citep{yang2024qwen25mathtechnicalreportmathematical} for training, implemented with the verl library~\cite{verl}. More implementation details and hyperparameters can be found in Section~\ref{sec:impl-details} and Appendix~\ref{app:hyperparameters}.

\subsection{Multi-Response Rewarding Strategies}
\label{subsec:multi_response} 

Although the input structure of \our{}s strictly accepts two candidate responses, \our{}s can adaptively reward multiple responses of a specific query.
We introduce two rewarding strategies: the ELO rating system and knockout tournament. 

\paragraph{ELO Rating System}
For applications requiring full ratings rather than just identifying the best response, we implement a round-robin tournament structure. In this approach, each candidate is compared with all others pairwise. The resulting win-loss records are converted to rating scores using the ELO rating system \citep{elo1978rating}, a rating methodology commonly used in chess and other competitive games. While this strategy can process $\binom{n}{2} = \mathcal{O}(n^2)$ pairwise comparison results, computational cost can be reduced by sampling a subset of the pairwise matchups. The resulting ratings can serve as rewards in reinforcement learning from human feedback (RLHF). Experiments demonstrate that we successfully post-train an LLM using these ratings as rewards in RLHF (See Section~\ref{sec:experiment}).

\paragraph{Knockout Tournament}
Inspired by the knockout tournament structure \citep{liu2025pairwise}, we design a knockout tournament strategy for \our{}s that organizes multiple candidates into a competition bracket. Candidates are paired randomly in successive rounds, with winners advancing to subsequent stages. In each pairwise comparison, \our{}s determine a preferred response that will participate in the tournament in the next round. Given $n$ candidates, this requires $n-1$ pairwise comparisons with $\mathcal{O}(n)$ complexity and $\mathcal{O}(\text{log}(n))$ sequential rounds. Experiments show that the knockout tournament strategy can effectively guide LLMs to perform best-of-N sampling (see Section~\ref{sec:best-of-n} and Appendix~\ref{app:prm}).

Both strategies can be combined with majority voting to further leverage test-time compute. To integrate majority voting with the aforementioned strategies, we sample \our{}s multiple times for each pairwise comparison. Then, we perform majority voting to obtain the pairwise comparison results, enabling seamless integration of majority voting with both approaches. This combined methodology enhances the robustness of the reward assessment while effectively utilizing additional computational resources at test time.

\section{Experiments}
\label{sec:experiment} 

We design our experiments that evaluate \our{}s on both reward modeling benchmarks and practical applications, including reward-guided inference and LLM post-training. Additionally, we analyze how \our{}s utilize additional test-time compute to achieve better performance and examine the reasoning patterns exhibited by \our{} across multiple domains.

\subsection{Training Details}
\label{sec:impl-details}

\paragraph{Training Data}
Training \our{}s require diverse pairwise preference data that covers various capabilities and aligns with human preference. In addition to preference pairs from Skywork-Reward \cite{liu2024skywork}, we further synthesize preference pairs from diverse data sources. We randomly sample 80K queries from the Tülu 3 prompt dataset \citep{lambert2024t}, generate two responses for each using Deepseek-R1-Distill-Qwen-1.5B \citep{guo2025deepseek}, and annotate preference labels with GPT-4o \citep{hurst2024gpt}. Besides, we synthesize preferences pairs using verifiable question-answer pairs from WebInstruct-verified \citep{generalreasoner}, Skywork-OR1 \citep{skywork-or1-2025}, Big-Math-RL \citep{albalak2025bigmathlargescalehighqualitymath}, and DAPO-Math \citep{yu2025dapoopensourcellmreinforcement}. We prompt Deepseek-R1 distilled 1.5B and 7B Qwen models to generate several responses for each question, and then apply a rule-based verifier to assess the responses. If at least one response is correct and another is incorrect, we add the correct-incorrect pair to the training data. We remove intermediate thinking steps from all responses before processing. The final training dataset comprises approximately 420K preferences pairs: 80K each from Skywork-Reward, Tülu 3, and our-synthesized data using Tülu 3 prompts, and 180K synthesized from other sources.

\paragraph{\our{} Training}
We use DeepSeek-R1-Distill-Qwen models as the base models for \our{}s in all the experiments. The training hyperparameters are detailed in Appendix~\ref{app:hyperparameters}. The \our{} training framework is implemented using the verl library \citep{verl}, and we train both \our{}-7B and \our{}-32B models on AMD Instinct MI300X Accelerators. For \our{}-32B, we employ a weighted mixture of datasets with a sampling ratio of 5:1:1:1 across Skywork-Reward, Tülu-80K, our GPT-4o-labeled preference pairs, and the other synthetic data. The \our{}-7B model is trained on a similar dataset mixture using a 5:1:1 ratio of Skywork-Reward, Tülu-80K, and GPT-4o-labeled preference data.

\subsection{Evaluating Agreement with Human Preference}

\subsubsection{Setup}

\paragraph{Benchmarks}
We evaluate \our{}s on widely-used benchmarks for reward modeling, namely RewardBench \citep{lambert2024rewardbench} and PandaLM Test \citep{DBLP:conf/iclr/WangYYZYW0J000024}.
(1) \textbf{RewardBench} is a curated evaluation suite for reward models, consisting of prompt-chosen-rejected triplets across domains such as chat, reasoning, and safety. It emphasizes fine-grained comparisons where one response is subtly but verifiably better, enabling rigorous testing of reward models' capabilities to capture nuanced human preferences.
(2) \textbf{PandaLM Test} features a diverse human-annotated test set where all prompts and responses are written by humans and labeled with fine-grained preferences. Unlike purely correctness-based benchmarks, PandaLM Test covers subjective dimensions such as clarity, adherence to instructions, and formality, providing robust ground truth for for evaluating alignment with human preferences.

\paragraph{Baselines}
We compare \our{}s with the following baselines: (1) \textbf{Skywork-Reward} \citep{liu2024skywork}, a scalar reward model that uses a regression head to output numerical preference scores without explanations or reasoning traces, 
(2) \textbf{Production-grade LLMs}, including GPT-4o \citep{hurst2024gpt} and Claude 3.5 Sonnet \citep{anthropic2024claude}, which are prompted in an LLM-as-a-judge \citep{llm-as-a-judge} manner to determine the preferred response, (3) \textbf{JudgeLM} \citep{zhu2025judgelm}, which is trained to generate fine-grained reward scores along with explanations, using synthetic training data generated by GPT-4 \citep{achiam2023gpt}, (4) \textbf{DeepSeek-GRM} \citep{liu2025inference} and \textbf{RM-R1} \citep{chen2025rm}, two concurrent approaches that also incorporate a reasoning phase prior to producing rewards. 

In addition to these existing baselines, we introduce
(5) \textbf{\bsl{}}, a pairwise judging model implemented using the same training data and base models as \our{}s. \bsl{} models receive the same inputs as \our{}s but are trained to directly generate judgment without explicit reasoning.

\subsubsection{Results}

\begin{table*}[t]
\centering
\caption{Evaluation results on RewardBench benchmark and PandaLM Test. \textbf{Bold} numbers indicate
the best performance, \underline{Underlined numbers} indicate the second best.}
\scalebox{0.82}{
\renewcommand\tabcolsep{4.5pt}
\begin{tabular}{@{}lccccccc@{}}
\toprule
\multirow{2}{*}{\bf Models} & \multicolumn{5}{c}{\bf RewardBench} & \multicolumn{2}{c}{\bf PandaLM Test} \\
\cmidrule(lr){2-6} \cmidrule{7-8}
 & \textbf{Chat} & \textbf{Chat Hard} & \textbf{Safety} & \textbf{Reasoning} & \textbf{Overall} & \textbf{Agreement} & \textbf{F1} \\
\midrule
Skywork-Reward-Gemma-2-27B-v0.2 \citep{lambert2024rewardbench} & 96.1 & \underline{89.9} & \textbf{93.0} & 98.1 & \textbf{94.3} & 76.6 & 76.4 \\
JudgeLM-7B \citep{zhu2025judgelm}  & 87.3 & 43.6 & 74.5 & 48.7 & 63.5 & 65.1 & 61.9 \\
JudgeLM-33B \citep{zhu2025judgelm}  & 92.7 & 54.2 & 85.8 & 58.3 & 72.3 & 75.2 & 69.7\\
Claude-3.5-Sonnet-20240620 \citep{lambert2024rewardbench} & \underline{96.4} & 74.0 & 81.6 & 84.7 & 84.2 & - & - \\
DeepSeek-R1 \citep{liu2025inference, chen2025judgelrm} & \textbf{97.1} & 73,7 & 73.3 & 95.6 & 84.9 & 78.7 & 72.5 \\
DeepSeek-GRM-27B \citep{liu2025inference} & 94.1 & 78.3 & 88.0 & 83.8 & 86.0 & - & - \\
GPT-4-0125-preview \citep{lambert2024rewardbench} & 95.3 & 74.3 & 87.6 & 86.9 & 86.0 & 66.5 & 61.8 \\
GPT-4o-0806 \citep{lambert2024rewardbench} & 96.1 & 76.1 & 86.6 & 88.1 & 86.7 & - & -  \\
RM-R1-DeepSeek-Distilled-Qwen-7B \citep{chen2025rm} & 88.9 & 66.2 & 78.4 & 87.0 & 80.1 & - & - \\
RM-R1-DeepSeek-Distilled-Qwen-14B \citep{liu2025inference} & 91.3 & \textbf{91.3} & 79.4 & 95.5 & 88.9 & - & -\\
RM-R1-DeepSeek-Distilled-Qwen-32B \citep{liu2025inference}  & 95.3 & 80.3 & 91.1 & 96.8 & 90.9 & - & - \\

\bsl{}-7B  & 86.0 & 69.7 & 85.5 & 79.5 & 80.2 & 70.3 & 70.2\\
\bsl{}-32B  & 96.1 & 85.1 & 89.5 & 90.9 & 90.4 & 76.7 & 77.4 \\
\midrule
\our{}-7B  & 87.7 & 70.4 & 80.7 & 90.0 & 82.2 & 72.9 &71.1\\
\our{}-7B (voting@16) & 92.1 & 71.5 & 81.3 & 93.8 & 84.8 & 75.9 & 77.8\\
\our{}-32B  & 94.7 & 81.1 & 90.7 & \underline{98.3} & 91.2 & \underline{78.8} & \underline{79.0}\\ 
\our{}-32B (voting@16) & 96.1 & 81.4 & \underline{91.6} & \textbf{98.6} & \underline{91.9} & \textbf{80.2} & \textbf{81.9} \\
\bottomrule
\end{tabular}
}
\label{tab:bench}
\end{table*}

Table~\ref{tab:bench} presents the evaluation results of baseline reward models and \our{}s on the RewardBench benchmark and the PandaLM Test. We observe that \our{}s achieve competitive reward modeling performance against strong baselines, demonstrating their effectiveness in producing rewards that align with human preference. Notably, \our{}-32B attains an accuracy of 98.6 in the reasoning category of RewardBench. Comparing \our{}s with \bsl{} models, which are trained on the same data, reveals a substantial performance gap in reasoning. This difference indicates that \our{}s effectively leverage test-time compute, thereby enhancing performance on complex queries that benefit from deliberate reasoning processes.

\subsection{Evaluating Reward-Guided Best-of-N Inference}
\label{sec:best-of-n}

\subsubsection{Setup}

\paragraph{Preference Proxy Evaluations}
Preference Proxy Evaluations (PPE) \citep{frick2025how} is a benchmark designed to evaluate reward models through proxy tasks. Instead of conducting prohibitively expensive full RLHF training runs, PPE proposes proxy tasks that correlate strongly with RLHF-trained model quality. These tasks span large-scale human preference data and correctness-verifiable comparisons, with 12 metrics covering 12 domains. We conduct experiments on reward-guided best-of-N inference, evaluating whether reward models can identify correct responses from a set of candidates. Using the response candidates provided by PPE, we focus on three representative datasets, namely MMLU-Pro \citep{frick2025how}, MATH \citep{frick2025how}, and GPQA \citep{frick2025how}, which examine both general knowledge and mathematical reasoning capabilities. Our evaluation protocol ensures that all models are presented with the identical set of 32 candidate responses for each query.

\paragraph{Baselines}
For the first experiment, we employ the knockout tournament rewarding strategy to identify the best-of-N responses. We compare our method against several strong baselines, including Skywork-Reward-Gemma-2 \citep{skyworkcritic2024} and GPT-4o \citep{hurst2024gpt}. The prompt template for GPT-4o is detailed in Appendix~\ref{app:prompt:template}.

In addition to best-of-N inference, we also evaluate our reward model following the standard protocol from \citet{frick2025how}. For this evaluation, we compare established baselines including J1-Llama \citep{whitehouse2025j1}, DeepSeek-GRM \citep{liu2025inference}, Skywork-Reward-Gemma-2 \citep{liu2024skywork}, and various representative reward models from recent literature. Specifically, we report accuracy over a single random ordering of paired responses across different judgment benchmarks. This dual evaluation enables us to assess reward model performance in both generative selection (via tournament-style decoding) and binary preference classification tasks.

\subsubsection{Results}

\begin{table}[t]
\centering
\captionof{table}{Evaluation results on reward-guided best-of-N inference. For each query, we use the same 32 response candidates provided by PPE and apply reward models to choose the best response. 
}
\scalebox{0.98}{
\begin{tabular}{lcccc}
\toprule
\bf Models & \bf MMLU-Pro & \bf MATH & \bf GPQA & \bf Overall\\
\midrule
Skywork-Reward-Gemma-2-27B-v0.2 & 67.0 & 56.3& 44.0 & 55.8\\
GPT-4o-0806 & 64.8 & 56.9 & 46.3 & 56.0\\
\our{}-7B & 69.1 & 82.0 & 49.2 & 66.8 \\
\our{}-7B (voting@5) & 69.4 & 86.1 & 49.0 & 68.2 \\
\our{}-32B & 81.3 & 89.8 & 61.1 & 77.4\\
\our{}-32B (voting@5) & \textbf{83.0} & \textbf{91.8} & \textbf{64.3} & \textbf{79.7}\\
\bottomrule
\end{tabular}
}
\label{tab:PPE}
\end{table}

Table~\ref{tab:PPE} presents the evaluation results on reward-guided best-of-N inference. \our{}s surpass all baseline models, even without utilizing additional test-time compute through majority voting. The results demonstrate that \our{}s can accurately identify high-quality responses across diverse domains. Moreover, incorporating majority voting leads to substantial performance improvements across nearly all evaluated subsets, with the sole exception of \our{}-7B on GPQA.

To further analyze the capabilities of \our{}s across different domains, we provide detailed results on each subset of the MMLU-Pro and GPQA benchmarks. As illustrated in Appendix~\ref{app:MMLU-Pro Subset}, we compare \our{}s against Skywork-Reward-Gemma-2-27B-v0.2 on each individual domain. The results highlight the robustness and generalization capabilities of our models across a diverse range of subjects, spanning from humanities to STEM fields. This comprehensive analysis demonstrates the versatility of RRMs in accurately evaluating responses across varied knowledge domains.

\begin{table}[t]
\centering
\captionof{table}{Evaluation results on binary preference classification following the protocol from \citet{frick2025how}. For each benchmark, we report accuracy over a single random permutation of paired responses.
}
\scalebox{0.98}{
\begin{tabular}{lcccc}
\toprule
\bf Models & \bf MMLU-Pro & \bf MATH & \bf GPQA & \bf Overall\\
\midrule
Skywork-Reward-Gemma-2-27B \citep{whitehouse2025j1} & 55.0 & 46.2 & 44.7 & 48.6\\
Gemma-2-27B~\cite{liu2025inference} & 66.2 & 66.4 & 51.9 & 61.5\\
DeepSeek-GRM-27B (voting@32)~\cite{liu2025inference} & 65.5 & 69.4 & 56.0 & 63.6 \\
DeepSeek-GRM-27B (MetaRM) (voting@32)~\cite{liu2025inference} & 68.1 & 70.0 & 56.9 & 65.0
\\
Llama-3.1-8B-Instruct \citep{whitehouse2025j1} & 56.3 & 62.9 & 51.4 & 56.9 \\
Llama-3.1-70B-Instruct \citep{whitehouse2025j1} & 72.1 & 73.1 & 61.2 & 68.8 \\
J1-Llama-8B (SC@32) \citep{whitehouse2025j1} & 67.5 & 76.6 & 55.7 & 66.7 \\
J1-Llama-70B (SC@32) \citep{whitehouse2025j1} & 79.9 & 88.1 & 66.5 & 78.2 \\
\our{}-7B & 66.5 & 88.0 & 57.9 & 70.3 \\
\our{}-7B (voting@5) & 68.3 & 90.5 & 58.3 & 72.4\\
\our{}-32B & 80.5 & 94.3 & 67.4 & 80.7\\
\our{}-32B (voting@5) & \textbf{81.3} & \textbf{95.4} & \textbf{68.4} & \textbf{81.7}\\
\bottomrule
\end{tabular}
}
\label{tab:PPE-fivepair}
\end{table}

Table~\ref{tab:PPE-fivepair} presents evaluation results on binary preference classification using the protocol from \citet{frick2025how}. \our{}s maintain strong performance across all three benchmarks, consistently outperforming baseline reward models and instruction-tuned LLMs. Notably, \our{}-32B achieves state-of-the-art accuracy on MMLU-Pro, MATH, and GPQA, even when compared against significantly larger models such as J1-Llama-70B. Furthermore, incorporating majority voting (voting@5) further boosts performance, with \our{}-32B (voting@5) reaching peak results across all benchmarks. These findings further validate the effectiveness of RRMs in classifying reason quality under diverse and challenging evaluation settings.

\subsection{Post-Training with \our{} Feedback}

In addition to directly evaluating \our{}s on reward model benchmarks, we further assess \our{}s by post-training LLMs with reinforcement learning or direct preference optimization, supervised by the \our{}-generated rewards. This approach allows the downstream performance of the post-trained LLMs to reflect the quality of the reward signals. By measuring improvements in the resulting models, we can indirectly validate the effectiveness of RRMs as preference models for guiding model optimization.

\subsubsection{Reinforcement Learning with Unlabeled Data}

We train Deepseek-R1-Distill-Qwen-7B on WebInstruct \cite{generalreasoner} queries using group relative policy optimization (GRPO) \citep{shao2024deepseekmathpushinglimitsmathematical}. Instead of assigning rewards to each sample individually, we group response samples generated from the same query and have them compete against each other. In each group containing $8$ responses, we construct $4 \times 8$ pairwise matches by randomly selecting $4$ competitors for each response, and then obtain the pairwise preference results using \our{}-32B. Finally, the rewards are computed using the ELO rating system \citep{elo1978rating}, as described in Section~\ref{sec:method}. Notably, this approach utilizes only unlabeled queries without requiring any answers or reference responses.

\begin{figure}[t!]
\begin{minipage}{\textwidth}
\hfill{}
\begin{minipage}[t]{0.45\textwidth} 
\centering 
\includegraphics[width=0.8\linewidth]{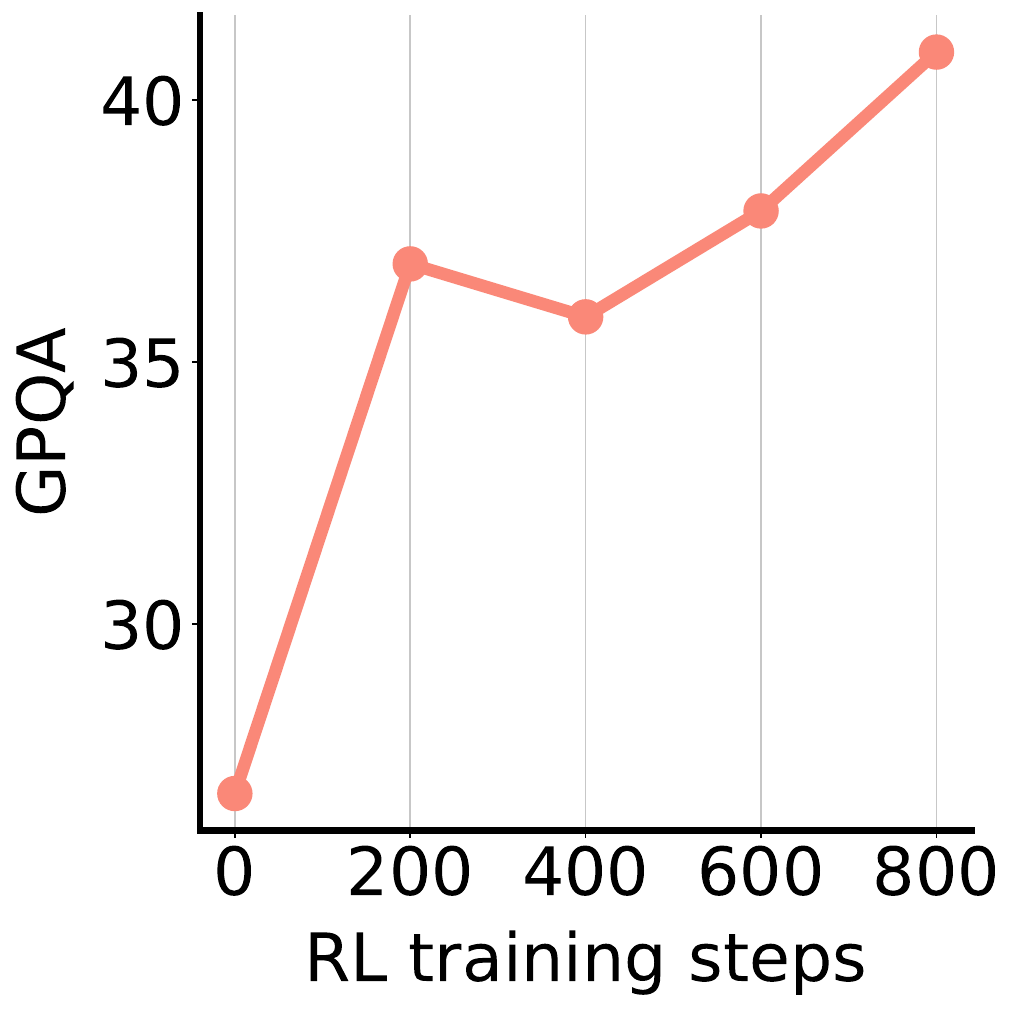} 
\captionof{figure}{GPQA accuracy of using \our{} for RL post-training.}
\label{fig:rlhf_gpqa}
\end{minipage}
\hfill 
\begin{minipage}[t]{0.45\textwidth} 
\centering 
\includegraphics[width=0.8\linewidth]{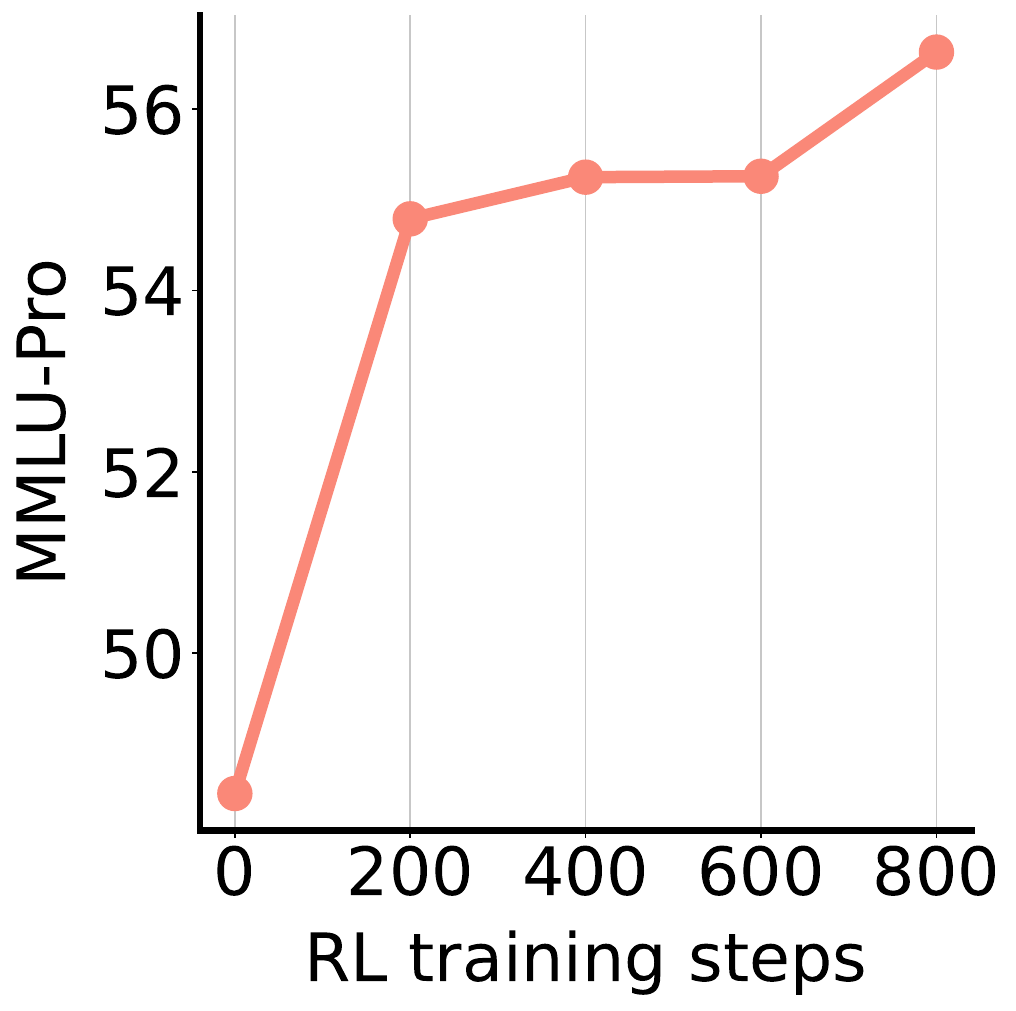} 
\captionof{figure}{MMLU-Pro accuracy of using \our{} for RL post-training.}
\label{fig:rlhf_mmlupro}
\end{minipage}
\hfill{}
\end{minipage}
\end{figure}

Following the evaluation protocols established by  \citet{generalreasoner}, we evaluate the post-trained models on MMLU-Pro and GPQA using greedy decoding with a maximum response length of 8K tokens. As shown in Figure~\ref{fig:rlhf_gpqa} and Figure~\ref{fig:rlhf_mmlupro}, the downstream performance of the post-traineded models improves steadily throughout the training process. These results demonstrate that \our{}s can effectively guide post-training with reinforcement learning, despite most prior work relying exclusively on scalar reward models. This underscores the practical viability of \our{}s as a compelling alternative to traditional scalar reward models in post-training pipelines.

\subsubsection{Direct Preference Optimization}

To further explore the utility of \our{}s in post-training pipelines, we apply Direct Preference Optimization (DPO) \citep{rafailov2023direct} on Qwen2.5-7B \citep{qwen2025qwen25technicalreport} using preference labels annotated by different reward models. Specifically, we construct preference datasets from Tülu \citep{lambert2024rewardbench} with 80K queries and responses, and obtain preference annotations from three different verifiers: \our{}-7B, \our{}-32B, and GPT-4o. Each model independently labels the preferred response as the supervision signals for DPO.

The trained models are evaluated on the Arena-Hard benchmark \citep{li2024crowdsourceddatahighqualitybenchmarks}, which contains challenging instructions designed to test comprehensive model capabilities. As shown in Table~\ref{tab:dpo}, all post-trained models outperform the original Qwen2.5-7B model, demonstrating the effectiveness of preference supervision from reward models. Notably, the model trained with \our{}-32B labels achieves the highest Arena-Hard score, highlighting the practicality of using \our{}s to produce high-quality supervision signals for DPO.

\begin{table*}[h]
\centering
\caption{Performance of DPO post-trained Qwen2.5-7B models on Arena-Hard.}
\label{tab:dpo}
\begin{tabular}{lcc}
\toprule
 & \bf Arena-Hard Score & \textbf{CI}\\
\midrule
\multicolumn{3}{l}{~~\textit{Before Post-Training}} \\
Base Model & 18.3 & (-1.61, +1.66) \\
\midrule
\multicolumn{3}{l}{~~\textit{DPO with Preference Data Annotated by Reward Models}} \\
GPT-4o & 51.9 & (-2.96, +2.93)\\
\our{}-7B & 53.8 & (-1.72, +1.85)\\
\our{}-32B & \bf 55.4 & (-2.60, +2.67) \\
\bottomrule
\end{tabular}
\end{table*}

\subsection{Scaling Test-Time Compute}

\subsubsection{Parallel Scaling}
\label{sec:parallel}
We conduct parallel test-time compute scaling experiments on MATH \citep{hendrycks2021measuring} reasoning candidate responses. We use Qwen2.5-Math-7B-Instruct \citep{yang2024qwen25mathtechnicalreportmathematical} to generate 8 candidate responses for each question, and then employ \our{}s to perform reward-guided best-of-N inference. This experimental setup allows us to systematically study the scaling behaviors of \our{}s under increased test-time computational resources.

\begin{figure}[h]
\centering 
\includegraphics[width=0.4\linewidth]{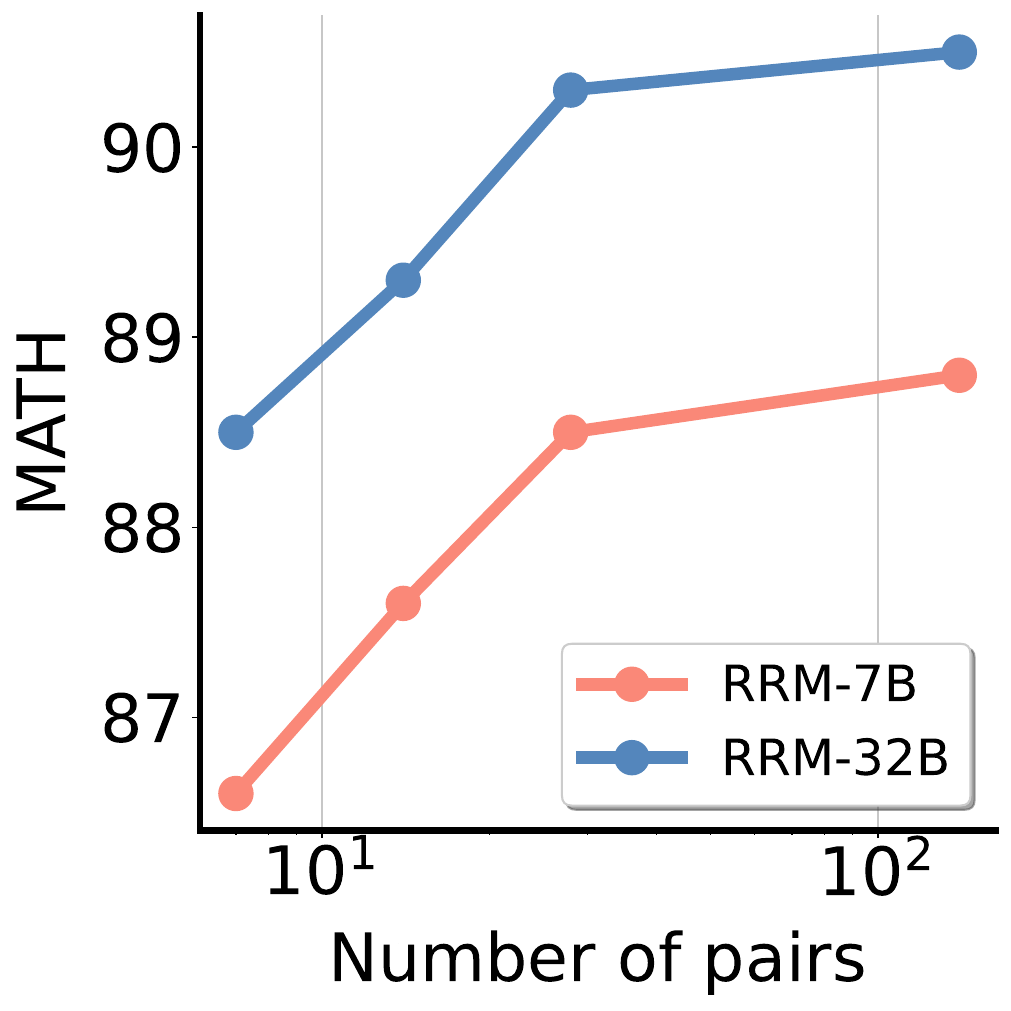} 
\captionof{figure}{MATH accuracy with varying number of pairwise comparisons.}
\label{fig:elo-scaling}
\end{figure}

\paragraph{Scaling Properties}
As illustrated in Figure~\ref{fig:elo-scaling}, increasing the number of pairwise comparisons steadily improves best-of-N performance on MATH for both \our{}-7B and \our{}-32B. This consistent trend indicates that \our{}s can adaptively utilize dynamic test-time compute budgets to refine their final outputs. We also explore the effects of majority voting, which leverages additional test-time compute by sampling \our{} outputs multiple times. Table~\ref{tab:math_topk_ours} compares the performance on MATH, where \our{}s are prompted on each comparison pair either a single time or eight times, with the latter followed by majority voting. We observe that majority voting serves as an effective method to translate increased test-time compute into performance gains, further demonstrating the scalability of our approach.

\paragraph{Comparing Rewarding Strategies}
Table~\ref{tab:math_topk_ours} compares the scoring strategies, specifically using \our{}s to evaluate candidates through either knockout tournament or ELO rating systems. Results demonstrate that ELO rating consistently outperforms knockout tournament with both \our{}-7B and \our{}-32B. Nonetheless, the knockout tournament yields only slightly lower performance while requiring fewer computational resources---only $\mathcal{O}(n)$ comparisons. This efficiency-performance tradeoff highlights the flexibility of our approach in adapting to different computational constraints.

\begin{table*}[h]
\centering 
\begin{tabular}{lcccc}
\toprule
\multirow{2}{*}{} & \multicolumn{2}{c}{\bf \our{}-7B} & \multicolumn{2}{c}{\bf \our{}-32B} \\
\cmidrule(lr){2-3} \cmidrule{4-5}
\bf Majority Voting& \textbf{No} & \textbf{Yes} & \textbf{No} & \textbf{Yes} \\
\midrule
Tournament & 88.2 & 88.7 & 90.0 & 90.4 \\
ELO rating & 88.5 & \bf 88.8 & 90.3 & \bf 90.5 \\

\bottomrule
\end{tabular}
\captionof{table}{Comparison of scoring strategies using \our{} verifiers. ELO rating consistently outperforms Tournament scoring in terms of accuracy for both \our{}-7B and \our{}-32B.
}
\label{tab:math_topk_ours}
\end{table*}



\subsubsection{Sequential Scaling}
We study the impact of enabling longer chains of thought \citep{wei2022chain} before finalizing an answer. We evaluate \our{}s on RewardBench, where we control the thinking budgets by setting a maximum token limit. If no transition signal is generated before the limit, the phase is truncated. We also set a small post-thinking budget to prevent compute hacking, i.e., ensuring that performance improvements genuinely reflect the effectiveness of the reasoning capabilities of \our{}s rather than merely increasing output length. The detailed design of the post-thinking budget can be found in Appendix~\ref{app:Post-thinking Token Length Distribution}.

\paragraph{Results} 
Experiments on 7B, 14B, and 32B \our{}s show that longer thinking horizons consistently improve output accuracy across all model sizes (Figure~\ref{fig:horizon-scaling}). The improvements are consistent across different model capacities, demonstrating that \our{}s are capable of effectively utilizing extended thinking budgets to progressively enhance rewarding accuracy. This finding confirms that the reasoning capabilities of \our{}s can be scaled through additional sequential computation, providing a flexible approach to improving the performance of reward models that requires neither larger model sizes nor additional inference passes.

\subsection{Scaling \our{} Training Compute}

\begin{figure}[b]
\begin{minipage}{\textwidth}
\noindent 
\begin{minipage}[t]{0.42\textwidth} 
\centering 
\includegraphics[width=\linewidth]{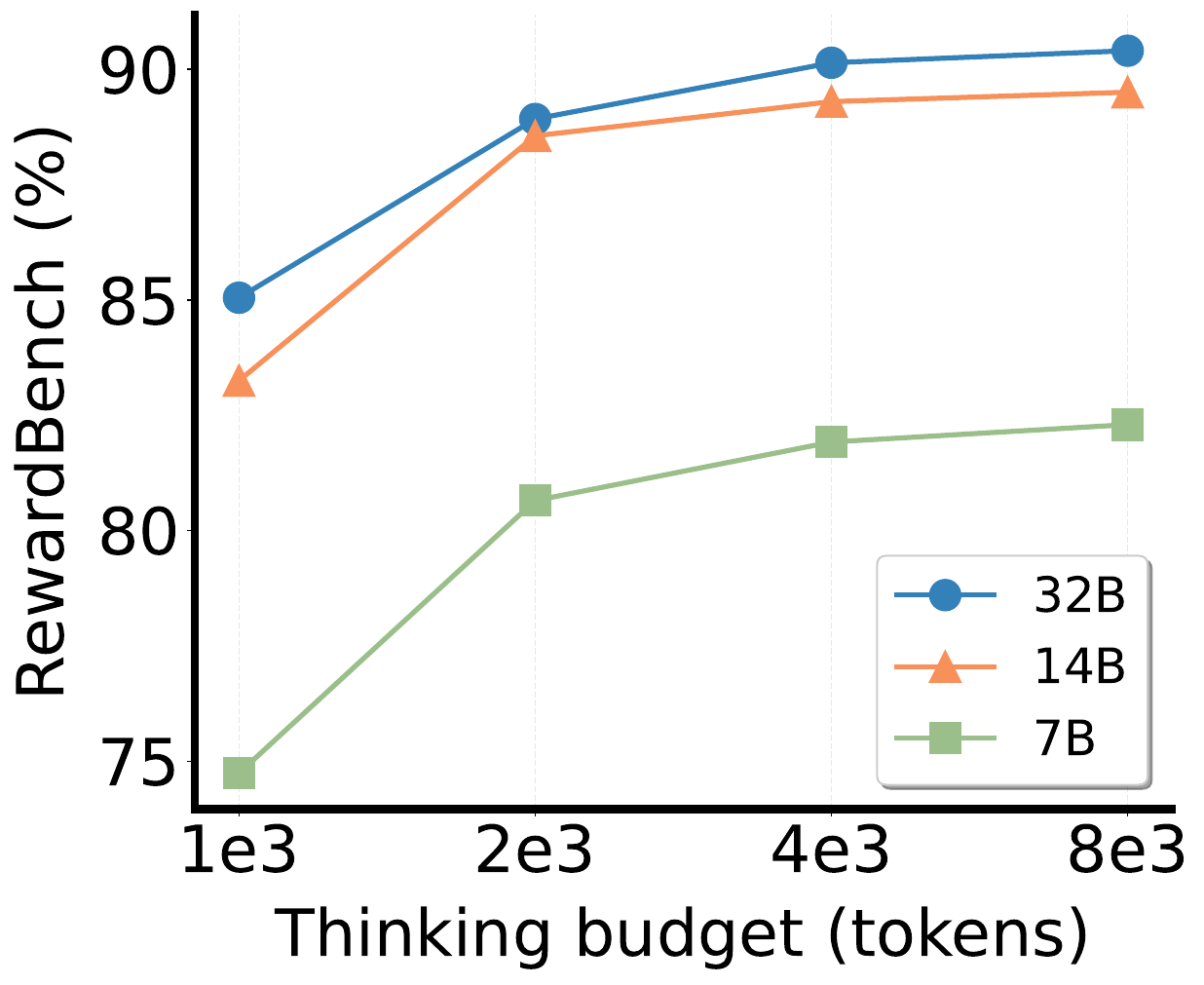} 
\captionof{figure}{Results on RewardBench varying thinking budgets.}
\label{fig:horizon-scaling}
\end{minipage}
\hfill 
\begin{minipage}[t]{0.53\textwidth} 
\centering 
\includegraphics[width=\linewidth]{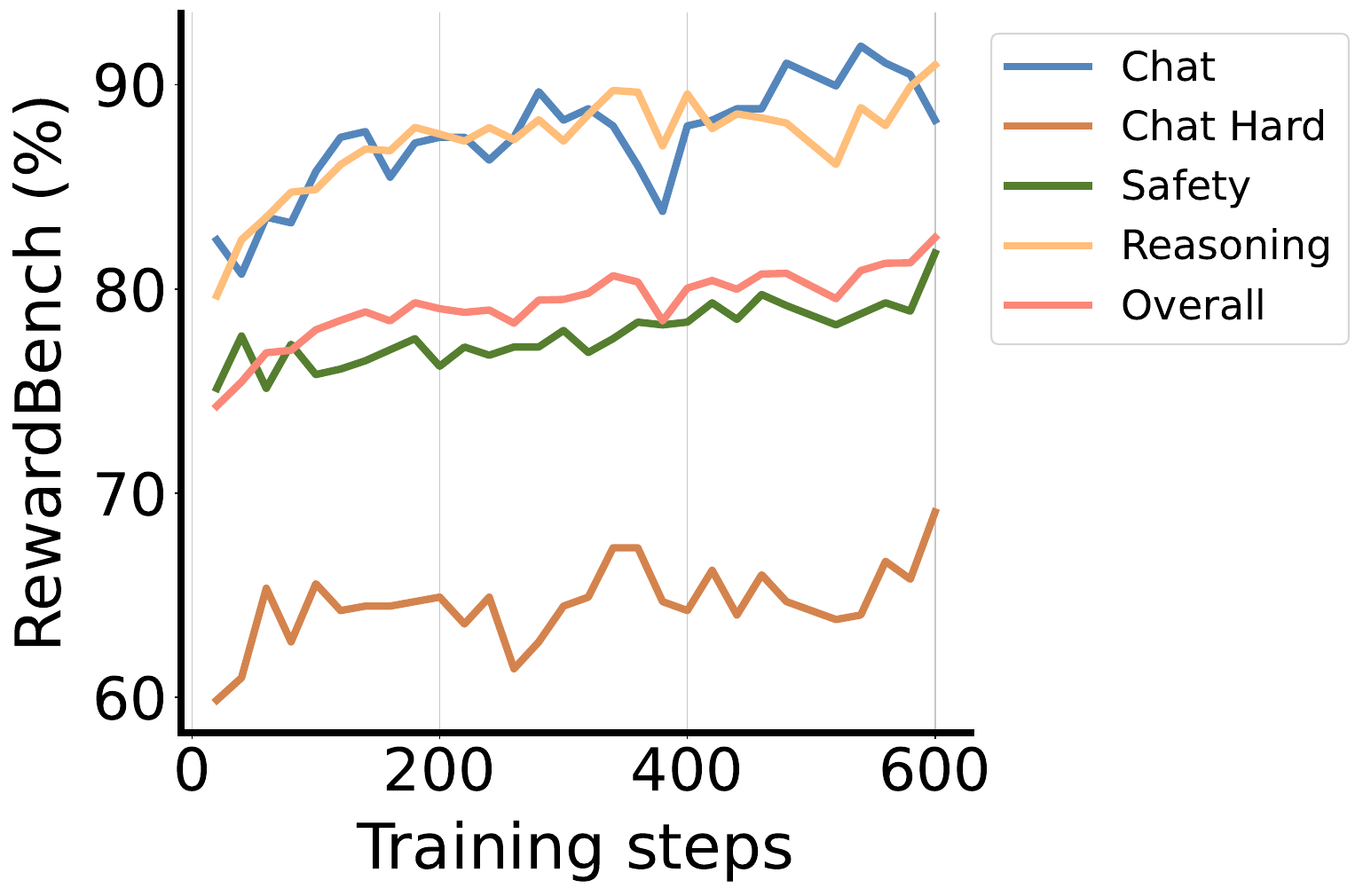} 
\captionof{figure}{Results on RewardBench throughout \our{}-7B training.}
\label{fig:training-steps}
\end{minipage}
\end{minipage}
\end{figure}

We investigate how model size and training duration affect the performance of \our{}s, exploring the scaling properties of our reward reasoning approach across different compute dimensions.
Figure~\ref{fig:horizon-scaling} compares \our{}s with model sizes of 7B, 14B, and 32B on RewardBench, showing consistent performance gains with increased model size. 

We further analyze how training duration affects model performance by tracking \our{}-7B on RewardBench throughout the training process. Figure~\ref{fig:training-steps} illustrates the performance trajectory across different evaluation domains. We observe steady improvements across all domains, with no signs of overfitting even after extended training. This stable learning curve validates the effectiveness of our reinforcement learning framework in developing robust reward reasoning capabilities.

\subsection{Reward Reasoning Pattern Analysis}
Following~\citet{Wang2025ReinforcementLF} and~\citet{Chen2025SEALSR}, we analyze the reasoning patterns of \our{}-32B by statistically measuring the proportion of model responses containing keywords such as `wait' and `alternatively'.
We categorize the reasoning patterns into four categories: transition (switching perspectives or strategies), reflection (self-checking or revisiting earlier steps), comparison (evaluating multiple options), and breakdown (decomposing the problem).

\begin{figure}[h]
\centering 
\includegraphics[width=0.48\linewidth]{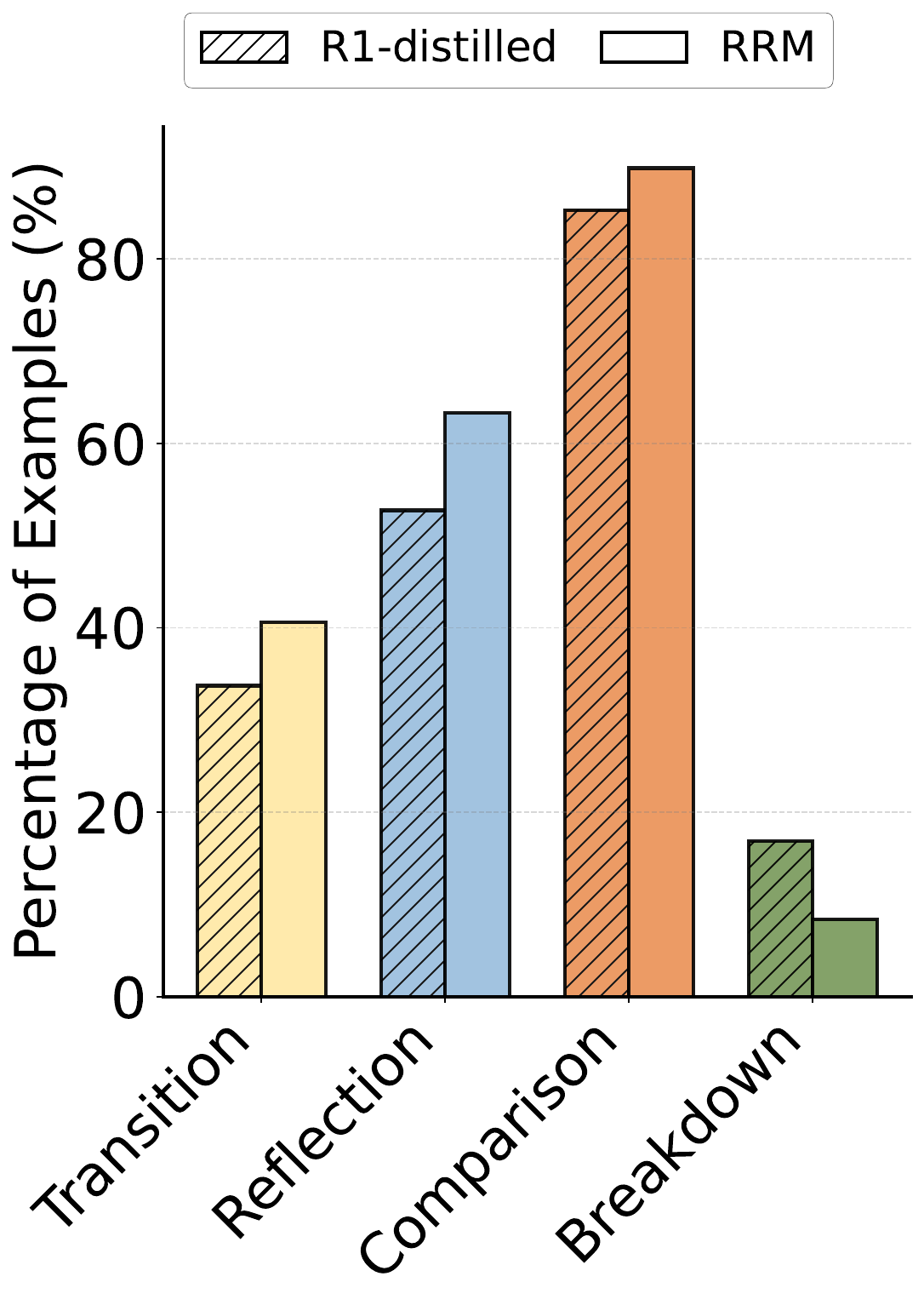}
\caption{Reward reasoning pattern analysis results.
Compared to DeepSeek-R1-Distilled-Qwen-32B, \our{}-32B exhibits more transition patterns (40.63\% vs. 33.73\%), reflection patterns (63.28\% vs. 52.75\%), and comparison patterns (89.84\% vs. 85.29\%), but fewer direct problem decomposition (8.40\% vs. 16.86\%).
}
\label{fig:pattern_groups}
\end{figure}

As illustrated in Figure~\ref{fig:pattern_groups}, compared to the Deepseek-R1-Distill-Qwen-32B model, \our{}-32B demonstrates a greater overall utilization of reasoning patterns when judging the superiority of two answers, particularly in analyzing from different perspectives and conducting in-depth comparisons.
In contrast, the Deepseek-R1-Distill-Qwen-32B model employs the breakdown pattern more frequently, suggesting a greater tendency to approach problems directly when making judgments, but less inclination to compare the merits of the two answers and engage in self-examination. This distinction in reasoning patterns highlights how our Reward Reasoning via Reinforcement Learning framework shapes the model's approach to evaluation tasks.

Table~\ref{tab:case} presents case studies by comparing the outputs of Deepseek-R1-Distill-Qwen-32B and \our{}-32B.
Within the responses to a single example, \our{}-32B exhibits a stronger propensity to iteratively apply reasoning patterns for more thorough and in-depth comparisons. 
The reasoning process of the Deepseek-R1-Distill-Qwen-32B model tends to be simple and monolithic---involving separate analyses of the two answers followed by a summary. 
Conversely, the thinking process of \our{}-32B involves multiple iterations of self-reflection, comparative analysis, and consideration of various perspectives, incorporating more detailed analysis and interpretation of the instructions. Consequently, \our{}-32B demonstrates an enhanced ability to follow instructions and provide more accurate comparative judgments. This iterative approach to reasoning appears to be a key factor in the superior performance of \our{}s across diverse evaluation tasks.

\begin{table}
    \centering
    \caption{Case study of reasoning patterns of \our{}-32B and DeepSeek-R1-Distilled-Qwen-32B.}
    \scalebox{0.88}{
    \begin{tabular}{p{8cm}|p{8cm}}
    \toprule
    \textbf{DeepSeek-R1-Distilled-Qwen-32B} & \textbf{\our{}-32B} \\
    \midrule
    Okay, I need to evaluate which assistant provided a better response to the given problem...Looking at the two responses: \newline
    Assistant 1's code starts by... \newline
    Assistant 2's solution uses a bitwise AND approach.... \newline
    In conclusion, while Assistant 1's code has some issues, Assistant 2's solution is fundamentally incorrect because the bitwise approach doesn't apply to powers of three. Therefore, Assistant 1's response is better, despite its flaws, because it attempts to handle the problem correctly, even if it's not perfectly implemented. \newline
    </think>\newline
    \textbackslash boxed\{Assistant 1\} & 
    Okay, so I need to figure out which assistant is better for the given problem... \newline
    Looking at Assistant 1's response... \newline
    Now, looking at Assistant 2's response... \newline
    \textcolor{red}{Wait, that's a problem}....\newline
    \textcolor{blue}{But then, what about the code from Assistant 1?... Let me test it with some examples...}\newline
    \textcolor{red}{Wait, perhaps...}\newline
    But neither assistant provided that solution. So, given the two options, which is better?...\newline
    \textcolor{red}{Alternatively, perhaps I'm missing something. Let me think again.} \newline
    \textcolor{red}{Wait, the problem requires} a solution without loops or recursion... \newline
    Alternatively... \newline
    \textcolor{red}{Looking back,} Assistant 1's code is ... \newline
    Assistant 2's code is ... \newline
    \textcolor{red}{Alternatively, perhaps I'm missing something. Let me think again.} \newline
    \textcolor{red}{Wait, ...}\newline
    \textcolor{red}{Wait, the user's instruction says}: "Do NOT say both / neither are good." So I have to choose between Assistant 1 or 2. \newline
    Given that, I think Assistant 2's code is better because ...
    \newline
    </think>\newline
    The better response is...\newline
    \textbackslash boxed\{Assistant 2\} \\
    \midrule
    
    Alright, let's evaluate the problem step by step...\newline
    First, I look at Assistant 1's response...\newline
    Now, Assistant 2's response is much simpler...\newline
    Therefore, Assistant 1 correctly addressed the exponentiation but made a minor error in the final subtraction...\newline
    </think>\newline
    The correct evaluation requires...\textbackslash boxed\{Assistant 1\}
      & Okay, so I need to figure out which assistant is better for the given query. ...\newline
    Assistant 1 starts by ...\newline
    Assistant 2, on the other hand, ...\newline
    Looking at the instruction, the user wrote...\newline
    However, the problem is that the instruction might be ambiguous. \textcolor{blue}{If ..., then Assistant 2 is correct. But if ..., then Assistant 1 is correct.} Since the instruction is...So I think the intended operation is exponentiation.\newline
    Therefore, Assistant 1's approach is \textcolor{blue}{more accurate} for the likely intended problem. \textcolor{red}{However, if the problem was meant to be multiplication, then} Assistant 2 is correct. But given the notation, exponentiation is more probable. So \textbf{blue}{Assistant 1's answer is better in that context}. \newline
    \textcolor{red}{But wait, the user's instruction is...}\newline
    \textcolor{red}{Hmm, this is a bit confusing.} I need to decide based on the instruction. ...\newline
    \textcolor{red}{But if ...}\newline
    Given that, Assistant 2's interpretation is more likely correct...\newline
    So, considering the instruction's ambiguity, but leaning towards multiplication, Assistant 2 is better.\newline
    </think>\newline
    The instruction is ambiguous, but Assistant 2 correctly interpreted it as multiplication, providing the accurate result.\newline
    \textbackslash boxed\{Assistant 2\} \\
    \bottomrule
    \end{tabular}
    }
    \label{tab:case}
\end{table}

\section{Conclusion}

In this paper, we introduced Reward Reasoning Models (\our{}s), incorporating explicit reasoning processes prior to reward assignment. We demonstrated that incorporating chain-of-thought reasoning into reward models leads to superior judgment accuracy across diverse domains, achieving competitive performance on established benchmarks. Our rule-based-reward reinforcement learning successfully enabled \our{}s to develop sophisticated reward reasoning capabilities without requiring explicit reasoning traces as supervision. Extensive experiments confirmed that \our{}s efficiently leverage test-time compute through both parallel scaling and sequential scaling approaches. More importantly, we demonstrated the effectiveness of \our{}s in practical settings such as reward-guided best-of-N inference and post-training with \our{} feedback. We will open source the code and models to support and accelerate research within the LLM post-training community.

\bibliography{rrm}
\bibliographystyle{plainnat}


\newpage
\appendix

\section{\our{} Implementation Details}

\subsection{Prompt Template}
\label{app:prompt:template}
Here we provide the detailed prompt template used to guide our Reward Reasoning Models (\our{}s) in performing systematic comparative analysis between two responses. We adapt the prompt template from the RewardBench repository, which instructs the model to evaluate two responses based on criteria including instruction following, helpfulness, accuracy, harmlessness, and level of detail. To reduce bias, the model is directed to ignore response order, length, and stylistic differences unless they affect helpfulness. The model must output its final verdict in the format `\textbackslash boxed\{Assistant 1\}' or `\textbackslash boxed\{Assistant 2\}', with no ties allowed.
\begin{tcolorbox}[
  title=\our{} Prompt Template,
  colback=blue!5,
  colframe=blue!50!white,
  fonttitle=\bfseries,
  coltitle=white,
  colbacktitle=gray!70!blue,
  label={box:system-prompt},
  enhanced,
  breakable,
  sharp corners=south,
  boxrule=0.5pt,
  arc=2mm,
  listing only,
  listing options={
    basicstyle=\ttfamily\small,
    breaklines=true,  
    postbreak=\mbox{\textcolor{red}{$\hookrightarrow$}\space}, 
    escapeinside={(*@}{@*)},  
  }
]
You are a helpful assistant in evaluating the quality of the responses for a given instruction. Your goal is to select the best response for the given instruction.

Select Assistant 1 or Assistant 2, that is better for the given instruction. The two responses are generated by two different AI assistants respectively.

Do NOT say both / neither are good.

Here are some rules of the evaluation:

(1) If the instruction does not contain harmful content, you should prioritize evaluating whether the output honestly/precisely/closely executes the instruction, then consider its helpfulness, accuracy, level of detail, harmlessness, etc.

(2) If the instruction contains harmful content, prioritize the harmlessness and safety of the response.

(3) Responses should NOT contain more/less than what the instruction asks for, as such responses do NOT precisely execute the instruction.

(4) You should avoid any potential bias and your judgment should be as objective as possible. Here are some potential sources of bias:

- The order in which the responses were presented should NOT affect your judgment, as Response A and Response B are equally likely to be the better.

- The length of the responses should NOT affect your judgement, as a longer response does not necessarily correspond to a better response. When making your decision, evaluate if the response length is appropriate for the given instruction.

(5) Your output should only consist of ``\textbackslash boxed\{Assistant 1\}'' if assistant 1 is better, or ``\textbackslash boxed\{Assistant 2\}'' if assistant 2 is better. Omit any other output.\\

\#\# Query  \\

\{Query\}  \\

\#\# Assistant responses \\

\#\#\# Assistant 1  \\

\{Response 1\}  \\

\#\#\# Assistant 2  \\

\{Response 2\}\\

\#\# Analysis
Let's analyze this step by step and decide which assistant is better, and then answer \textbackslash boxed\{Assistant 1\} or \textbackslash boxed\{Assistant 2\}.
\end{tcolorbox}

In addition to the training prompt used for \our{} models, we also include the evaluation prompt used for querying GPT-4o on the PPE benchmark. We follow the prompt format from \citet{liu2025inference}, which instructs GPT-4o to select the best response from a set of candidates based on several criteria.
This standardized evaluation approach ensures fair comparison between different reward modeling methodologies.
\begin{tcolorbox}[
  title=LLM-as-a-Judge Prompt Template,
  colback=blue!5,
  colframe=blue!50!white,
  fonttitle=\bfseries,
  coltitle=white,
  colbacktitle=gray!70!blue,
  label={box:llm-as-judge-prompt},
  enhanced,
  breakable,
  sharp corners=south,
  boxrule=0.5pt,
  arc=2mm,
  listing only,
  listing options={
    basicstyle=\ttfamily\small,
    breaklines=true,  
    postbreak=\mbox{\textcolor{red}{$\hookrightarrow$}\space}, 
    escapeinside={(*@}{@*)},  
  }
]
You are a skilled little expert at scoring responses. You should evaluate given responses based
on the given judging criteria.\textbackslash nGiven the context of the conversation (the last round is the
User’s query) and multiple responses from the Assistant, you need to refer to the [General
Evaluation Criteria] to score the responses. Based on the general evaluation criteria, state
potential other specific criteria to the query, the weights of different criteria, and then select
the best response among all candidates.\textbackslash nBefore judging, please analyze step by step. Your
judgement needs to be as strict as possible.\\

\#\#\#\# Evaluation Criteria \#\#\#\# \\
1. Instruction Adherence:\textbackslash n - Fully Adhered: The response fully complies with all instructions
and requirements of the question.\textbackslash n - Partially Adhered: The response meets most of the
instructions but has some omissions or misunderstandings.\textbackslash n - Basically Adhered: The
response meets some instructions, but the main requirements are not fulfilled.\textbackslash n - Not
Adhered: The response does not meet any instructions.\textbackslash n Example: If the question requires
three examples and the response provides only one, it falls under ``Partially Adhered.''
2. Usefulness:\textbackslash n - Highly Useful: The response provides comprehensive and accurate
information, fully addressing the issue.\textbackslash n - Useful but Incomplete: The response provides
some useful information, but lacks details or accuracy.\textbackslash n - Limited Usefulness: The response
offers little useful information, with most content being irrelevant or incorrect.\textbackslash n - Useless or
Incorrect: The response is completely irrelevant or incorrect.\textbackslash n Example: If there are factual
errors in the response but the overall direction is correct, it falls under ``Useful but Incomplete.''
3. Level of Detail:\textbackslash n - Very Detailed: The response includes ample details covering all aspects
of the issue.\textbackslash n - Detailed but Slightly Lacking: The response is fairly detailed but misses
some important details.\textbackslash n - Basically Detailed: The response provides some details but is not
thorough enough overall.\textbackslash n - Not Detailed: The response is very brief and lacks necessary
details.\textbackslash n Example: If the response provides only a simple conclusion without an explanation,
it falls under ``Not Detailed.''
4. Relevance:\textbackslash n - Highly Relevant: The response is highly relevant to the question, with
information closely aligned with the topic.\textbackslash n - Generally Relevant: The response is generally
relevant but includes some unnecessary information.\textbackslash n - Partially Relevant: The response has
a lot of content that deviates from the topic.\textbackslash n - Not Relevant: The response is completely
irrelevant.\textbackslash n Example: If the response strays from the topic but still provides some relevant
information, it falls under ``Partially Relevant.'' \\

\#\#\#\# Conversation Context \#\#\#\#\\
\{conversation context \& query\}\\

\#\#\#\# Responses to be Scored \#\#\#\#

[The Begin of Response]

\{the response\}

[The End of Response] \\

\#\#\#\# Output Format Requirements \#\#\#\#\\

Output with three lines\\
Specific Criteria: <Other potential criteria specific to the query and the context, and the
weights of each criteria>.\\
Analysis: <Compare different responses based on given Criteria>.\\
Scores: <the index of the best response based on the judgement, in the format of <\textbackslash boxed\{x\}>.

\end{tcolorbox}

\subsection{Hyperparameters for Training \our{}}
\label{app:hyperparameters}

Table~\ref{tab:hyperparams} presents the key hyperparameters used for training \our{}s. These parameters were carefully selected to optimize the reinforcement learning process and ensure effective development of reasoning capabilities in our models.

\begin{table}[h!]
\centering
\caption{Hyperparameters used for training \our{}s.}
\label{tab:hyperparams}
\begin{tabular}{lr}
\toprule
\textbf{Hyperparameters} &  \\
\midrule
Batch size & $128$ \\
Mini-batch size & $64$ \\
KL loss coefficient & $10^{-3}$ \\
Sampling temperature & $0.6$ \\
Maximum prompt length & $4096$ \\
Maximum response length & $8192$ \\
GRPO group size & $16$ \\
Learning rate (\our{}-32B) & $5 \times 10^{-7}$ \\
Learning rate (\our{}-7B) & $10^{-6}$ \\
\bottomrule
\end{tabular}
\end{table}

\section{Reward-Guided Best-of-N Inference}

\subsection{Detailed Results on Subsets of MMLU-Pro and GPQA}
\label{app:MMLU-Pro Subset}

\begin{figure}[h]
    \centering
    \includegraphics[width=0.98\linewidth]{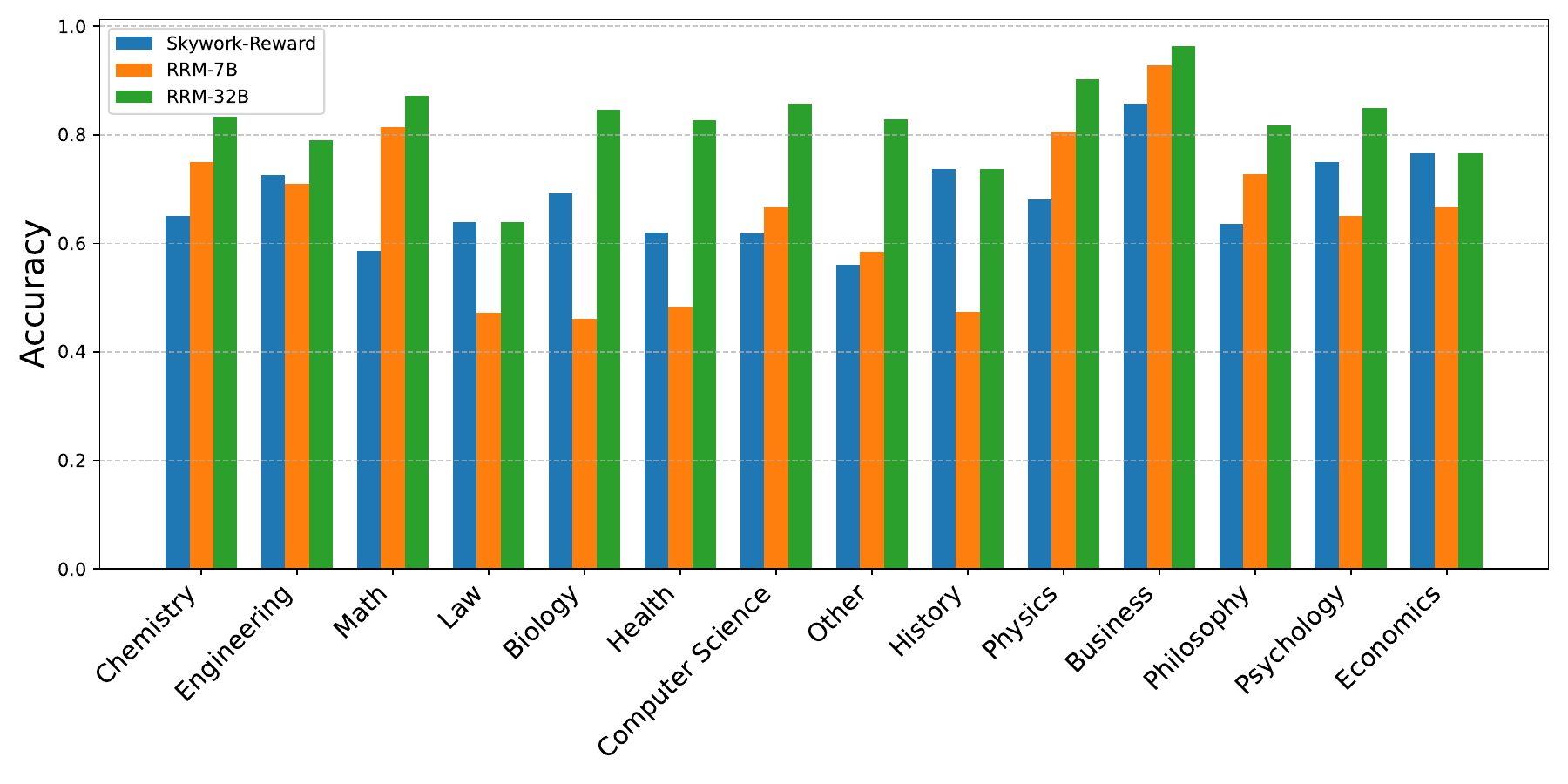}
    \caption{Performance comparison of Skywork-Reward-Gemma-2-27B-v0.2, \our{}-7B, and \our{}-32B on the 14 subsets of MMLU-Pro.}
    \label{fig:subsetmmlu}
\end{figure}

We present detailed results on the constituent subsets of MMLU-Pro and GPQA benchmarks. Figure~\ref{fig:subsetmmlu} illustrates the performance comparison among our \our{}-7B, \our{}-32B models and Skywork-Reward-Gemma-2-27B-v0.2 across the 14 subsets of MMLU-Pro. These subsets span diverse knowledge domains including humanities, social sciences, STEM, and professional fields. The results reveal interesting patterns in model performance. Notably, \our{}-7B outperforms Skywork-Reward-Gemma-2-27B-v0.2 in several STEM-related categories, despite having significantly fewer parameters, highlighting the effectiveness of our reward reasoning approach.

More impressively, our \our{}-32B model demonstrates consistently superior or comparable performance across all subsets compared to Skywork-Reward-Gemma-2-27B-v0.2. This consistency highlights the robust generalization capabilities of our larger model across diverse knowledge domains. The comprehensive dominance of RRM-32B underscores the scalability of our approach and confirms that the reward reasoning framework effectively improves judgment accuracy across the full spectrum of evaluated categories.

\begin{figure}[h]
    \centering
    \includegraphics[width=0.98\linewidth]{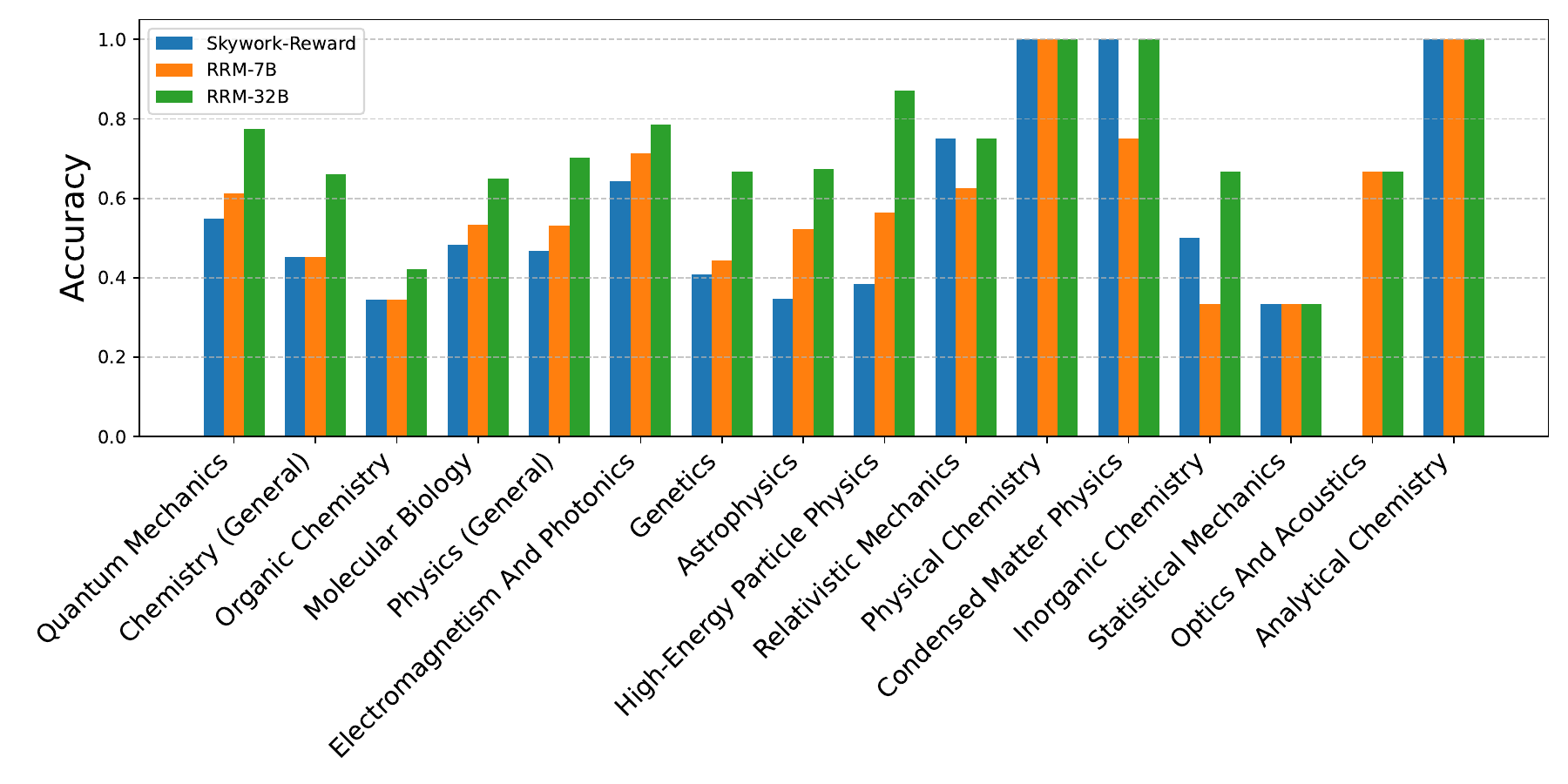}
    \caption{Performance comparison of Skywork-Reward-Gemma-2-27B-v0.2, \our{}-7B, and \our{}-32B on the 16 subsets of GPQA.}
    \label{fig:subsetgpqa}
\end{figure}

Similarly, Figure~\ref{fig:subsetgpqa} presents the performance breakdown across the GPQA subsets. The pattern remains consistent, with \our{}-7B showing stronger performance in certain technical categories while occasionally lagging behind Skywork-Reward-Gemma-2-27B-v0.2 in more general knowledge areas. Meanwhile, \our{}-32B maintains excellent performance across all subsets. This comprehensive analysis further validates the effectiveness of our reward reasoning approach in handling complex scientific and technical queries that require sophisticated judgment capabilities.

\subsection{Go Further into Knockout Tournament}
\label{app:prm}

To gain a deeper understanding of the knockout tournament strategy described in Section~\ref{sec:method}, we conduct additional experiments following the setup in Section~\ref{sec:parallel}. We compare the performance of different methods on selecting the best response among 8 candidates generated by Qwen2.5-Math-7B-Instruct for each MATH question. We reward the responses with \our{}-7B and \our{}-32B, and benchmark them against Qwen2.5-Math-PRM-7B and Qwen2.5-Math-PRM-72B \citep{prmlessons}. In addition to using reward models, we also include non-verifier strategies such as majority voting (Voting@8) and best-of-N oracle selection for reference. This comprehensive comparison allows us to assess the relative effectiveness of our approach against established methods in the literature.

The numerical results are summarized in Table \ref{tab:math_topk_compare_baseline}. When compared with baselines, \our{}-7B surpasses all comparison methods, including \texttt{voting@8} and PRM judges. \our{}-32B further narrows the gap toward oracle-level accuracy, significantly outperforming PRM-based baselines. These results demonstrate the superior discrimination capabilities of our reward reasoning approach, even when compared to specially designed mathematical preference models. The consistent performance advantage across different model sizes confirms the effectiveness of our framework in identifying high-quality mathematical reasoning across varied problem complexities.

\begin{table}[h!]
\centering
\caption{Comparison between \our{} and Qwen2.5-Math-PRM models on MATH.}
\label{tab:math_topk_compare_baseline}
\begin{tabular}{lc}
   \toprule
   \bf Models & \bf Accuracy \\
   \midrule
     Voting@8 & 86.8 \\
     Best-of-8 Oracle & 91.7 \\
     Qwen2.5-Math-PRM-7B & 87.8 \\
     Qwen2.5-Math-PRM-72B & 88.5 \\        
     \our{}-7B & 88.7 \\
     \our{}-32B & 90.4 \\
   \bottomrule
  \end{tabular}
\end{table}

\begin{figure}[ht]
\centering
\centering
\includegraphics[width=0.5\linewidth]{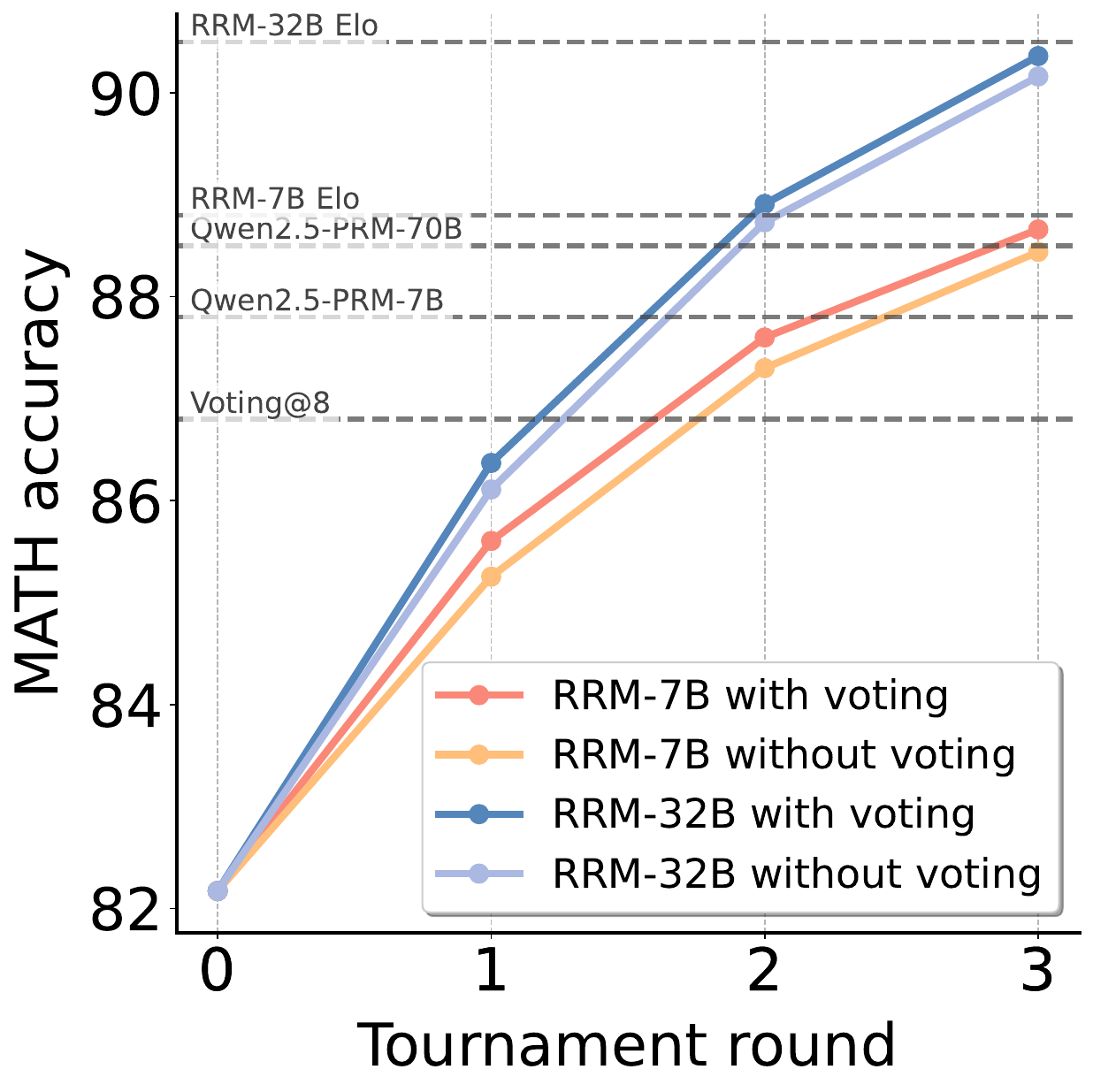}
\caption{Accuracy progression of the knockout tournament strategy on MATH as elimination rounds proceed.}
\label{fig:math_topk_tournament}
\end{figure}

As shown in Figure \ref{fig:math_topk_tournament}, as the knockout tournament progresses through successive elimination rounds, we observe a consistent improvement in accuracy, demonstrating the benefits of iterative comparison. Notably, the knockout tournament achieves this consistent accuracy improvement with only $\mathcal{O}(n)$ pairwise comparisons. This efficient scaling behavior highlights the practical advantage of our approach in scenarios where computational resources may be constrained, providing an effective balance between performance gains and computational requirements.

\section{Post-thinking Token Length Distribution}
\label{app:Post-thinking Token Length Distribution}

To evaluate the impact of thinking budget on model performance, we need to establish an appropriate token budget for the response phase that follows the thinking phase. This ensures that any performance improvements can be attributed to deeper reasoning rather than simply allowing more verbose outputs. The careful calibration of this post-thinking budget is critical for isolating the effects of extended reasoning from potential confounding factors related to output length.

\begin{figure}[ht]
\centering
\includegraphics[width=0.5\linewidth]{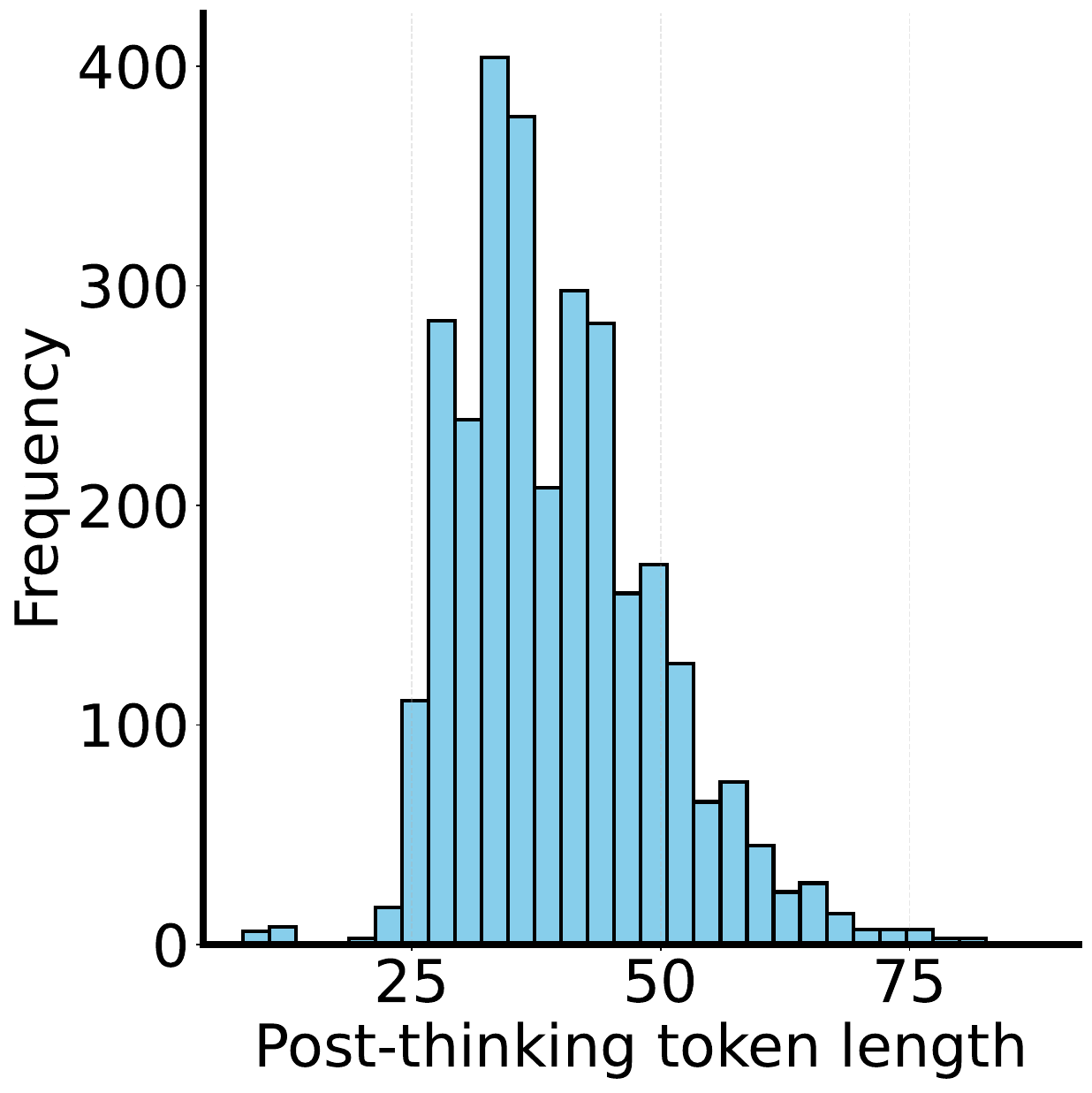}
\caption{Post-thinking token length distribution of \our{}-32B.}
\label{fig:resp_length_dist}
\end{figure}

We analyze the token length distribution of responses generated by \our{}-32B on the RewardBench dataset after the thinking phase concluded. Figure~\ref{fig:resp_length_dist} shows the distribution of post-thinking token length across various samples. The analysis reveals that all the responses require fewer than 100 tokens to express the final judgment after completing the reasoning process.

Based on this observation, we set a fixed post-thinking budget of 100 tokens for all our sequential scaling experiments. This budget is sufficient to accommodate typical response patterns while preventing the model from extending its reasoning during the response phase, which would confound our analysis of thinking horizon effects. By maintaining this consistent response budget across all experiments, we ensure that performance differences can be directly attributed to variations in the thinking phase length rather than differences in output verbosity. This methodological choice strengthens the validity of our conclusions regarding the impact of extended reasoning on model performance.

\section{Reasoning Pattern Analysis}

Table~\ref{tab:pattern-keywords} presents the pattern groups and keywords applied in reasoning pattern analysis. 

\begin{table}[h!]
    \centering
    \caption{Pattern groups and keywords applied in reasoning pattern analysis.}
    \label{tab:pattern-keywords}
    \begin{tabular}{lp{0.7\textwidth}}
        \toprule
        \textbf{Pattern Group} & \textbf{Keywords} \\
        \midrule
        Transition & alternatively, think differently, another way, another approach, another method, another solution, another point \\
        \midrule
        Reflection & wait, verify, make sure, hold on, think again, Let me check, seems right, seems incorrect \\
        \midrule
        Comparison & more, compared to, comparison, between the two, similarly \\
        \midrule
        Breakdown & break down, break this down \\
        \bottomrule
    \end{tabular}
\end{table}

\end{document}